\documentclass[pdflatex,sn-apa]{sn-jnl}

\usepackage[english]{babel}
\usepackage{graphicx}
\usepackage{array}
\usepackage{makecell}
\usepackage{multirow}
\usepackage[table]{xcolor}
\usepackage{float}
\usepackage{placeins}
\usepackage{tabularx}
\usepackage{booktabs}
\usepackage{amsmath}
\usepackage{amssymb}
\usepackage{enumitem}
\usepackage{microtype}
\begin{document}

\title[Temporal Dropout Risk in Learning Analytics]{A Unified Survival Benchmark for Temporal Dropout Risk Prediction in Learning Analytics}

\author*[1]{\fnm{Rafael} \sur{da Silva}}\email{rafael.dasilva@eastern.edu}

\author[1]{\fnm{Jeff} \sur{Eicher}}\email{jeff.eicher@eastern.edu}

\author[1]{\fnm{Gregory} \sur{Longo}}\email{gregory.longo@eastern.edu}

\affil*[1]{\orgdiv{Applied Data Science Program},
           \orgname{Eastern University},
           \orgaddress{\city{St.~Davids}, \state{PA}, \country{USA}}}

\abstract{Although dropout remains a salient challenge in Learning Analytics, head-to-head model comparisons often rest on mismatched evaluation setups that emphasize ranking ability more than time-aware interpretability or calibration quality. Here we present a survival-oriented benchmark for temporal dropout risk modelling using the Open University Learning Analytics Dataset (OULAD). Two arms are compared: Family~A: Dynamic Weekly, with models in person-period representation, and Family~B: Static Early-Window, with an expanded roster of families: tree-based survival, parametric, and neural models. The evaluation protocol integrates four analytical layers: predictive performance, ablation, explainability, and calibration. Results are reported within each family separately, because a single numerical cross-family ranking would conflate genuine model differences with artifacts of temporal representation, to which survival metrics are known to be sensitive. Within Family~B, Random Survival Forest showed the highest point estimates for time-dependent concordance and the lowest Brier scores across all three horizons; within Family~A, Poisson Piecewise-Exponential showed the lowest point estimate for integrated Brier score within a tight five-model cluster. No-refit bootstrap resampling qualifies these positions as directional signals, not claims of strict superiority. Ablation and explainability analyses converged, across all models, on a shared finding: the dominant predictive signal was not primarily demographic or structural, but temporal and behavioral. Calibration corroborated this pattern in the better-discriminating models, except for XGBoost AFT, the sole outlier (analyzed in the Discussion). These results support unified, multi-dimensional benchmarking in Learning Analytics and situate dropout risk as a temporal-behavioral process rather than a function of static background attributes.}


\maketitle

\section{Introduction}

Student dropout poses a sustained challenge for higher education and for Learning Analytics as a field, driven not only by institutional and social costs, but by the difficulty of identifying risk patterns in time to support retention and persistence \citep{Nagy2023InterpretableDropout,SandovalPalis2020EarlyDropout}. Although this problem has attracted substantial research attention, a core tension persists: predictive models tend to be evaluated in terms of \textit{who} is at risk, with less precision about \textit{when} that risk intensifies across the academic trajectory \citep{Adnan2021EarlyIntervention,Seo2024DropoutRisk}.

The temporal dimension matters because dropout risk is not a static condition; it unfolds as engagement, activity, and academic progression accumulate week by week \citep{Mubarak2020InteractionLogs,Baneres2023IJETHE}. Yet the literature on dropout prediction still exhibits gaps at three interconnected levels. Many studies compare models under heterogeneous evaluation protocols, making it difficult to draw firm conclusions about differences across models \citep{Oqaidi2022DropoutModel,Coussement2020LogitLeaf}. Even when comparative performance is strong in terms of discrimination, evaluations do not always incorporate dimensions essential for educational use: calibration and interpretability chief among them; calibration, in particular, refers to the numerical coherence between predicted risk probabilities and observed dropout rates \citep{Nagy2023InterpretableDropout}. And the literature does not consistently translate performance differences into a substantive account of what kind of signal structures dropout risk: whether static student attributes or temporal and behavioral signals accumulated throughout the course \citep{Seo2024DropoutRisk,Zhidkikh2024CS1}.

This paper responds to those gaps through a survival-oriented benchmark for temporal dropout risk modelling in Learning Analytics. The empirical basis is the Open University Learning Analytics Dataset (OULAD), a widely used resource in Learning Analytics and Educational Data Mining \citep{Kuzilek2017OULAD}. Its choice is deliberate rather than conventional: as the most-reused empirical testbed in the field, OULAD acts here as a standardized substrate whose familiarity allows performance differences to be read against a known baseline. The paper's contribution is not a new state-of-the-art score on this dataset; it is a demonstration that the same data yield different conclusions depending on how models are evaluated, which representational families they are organized into, and whether the evaluation protocol is unified across them. Data novelty is not the claim; evaluation architecture is. Under a unified evaluation framework---a single evaluation protocol (shared metrics, preprocessing, tuning, and infrastructure) applied across model families while preserving each family's methodological scope, not a merging or pooling of multiple datasets---two families are compared: Family~A: Dynamic Weekly, with models in person-period representation, and Family~B: Static Early-Window, with an expanded roster of families covering tree-based survival, parametric, and neural approaches. The benchmark is designed to support within-family comparison under consistent evaluation protocols, making explicit the representational distinction between dynamic weekly and early-window approaches, under which a single numerical cross-family ranking would conflate model differences with artifacts of temporal representation. Crucially, the two families are two representations of the \emph{same} underlying engagement signal, drawn from the same VLE logs; running them as parallel arms turns the representation choice into an experimental control, so any finding that holds across both families is robust to representation rather than an artifact of one encoding. The division is a control, not an arbitrary partition.

The benchmark is organized around four complementary dimensions. Predictive performance is evaluated through survival-oriented metrics under shared benchmark horizons: the Integrated Brier Score (IBS; a mean squared error integrated over the survival time axis) quantifies global probabilistic error, while time-dependent concordance (C-index) measures ordinal discriminative ability across the trajectory. An ablation analysis then assesses the extent to which performance depends on static feature blocks versus temporal and behavioral signals. An explainability layer identifies each family's dominant predictive drivers and compares the relative importance of feature blocks. A calibration layer is included because it is not enough for a model to rank cases well; predicted risks must also remain numerically coherent with observed outcomes, particularly when those estimates may inform support allocation and intervention decisions \citep{Seo2024DropoutRisk,VanCalster2019CalibrationAchilles}. Each layer earns its place: predictive performance asks \emph{how well} models rank and predict; ablation asks \emph{where} the signal comes from (static vs.\ temporal blocks); explainability asks \emph{which} block dominates; and calibration asks whether the predicted probabilities are \emph{numerically honest} enough to drive support decisions. The four are read jointly, not as interchangeable summaries.

Results are reported within each family separately, because a single numerical cross-family ranking would conflate genuine model differences with artifacts of temporal representation, to which survival metrics are known to be sensitive. Within Family~B, Random Survival Forest showed the highest point estimates for time-dependent concordance and the lowest Brier scores across all three horizons; within Family~A, Poisson Piecewise-Exponential showed the lowest point estimate for integrated Brier score within a tight five-model cluster. No-refit bootstrap sampling variability qualifies these positions as directional signals rather than absolute superiority (sampling variance only; see \S\ref{subsec:bootstrap_scope}). More substantively, ablation and explainability converged on a consistent finding across all models: the dominant predictive signal was not primarily demographic or structural, but temporal and behavioral \citep{Marcolino2025MoodleLogs}. Calibration corroborated this pattern in the better-discriminating models, except for XGBoost AFT, the sole outlier (analyzed in the Discussion). This convergence is what justifies maintaining two families rather than one: temporal-behavioral dominance appears in both representations and across linear, tree-based, and neural architectures---evidence no single family could provide.

The study contributes on four fronts: a unified evaluation framework that shares metrics, protocols, and infrastructure across two representational arms while preserving the methodological separation that makes within-family comparisons valid; an argument that model comparison must extend beyond discrimination to calibration and interpretation; empirical evidence of temporal-behavioral signal dominance over static covariates; and a substantive reading of dropout risk as a structured temporal-behavioral process \citep{Oqaidi2022DropoutModel,Marcolino2025MoodleLogs,Mubarak2020InteractionLogs}.

Three research questions guide the article. \textbf{RQ1:} How do different survival-oriented models compare, under a unified evaluation protocol, in temporal dropout risk prediction? \textbf{RQ2:} To what extent does model performance depend on temporal and behavioral signals relative to static covariates? \textbf{RQ3:} How do explainability and calibration patterns help interpret, from a Learning Analytics perspective, what structures predicted dropout risk?

\section{Literature Review}

Dropout prediction occupies an established position within Learning Analytics, particularly in research on retention, persistence, and early identification of at-risk learners. Recent reviews of Predictive Learning Analytics in higher education confirm that retention and dropout are among the most frequently modelled outcomes, marking this as a well-developed line of inquiry \citep{Sghir2022PLAReview}. This research agenda is closely tied to an institutional interest in using analytics not merely to describe learner behaviour, but to support timely decisions about intervention and student support \citep{Sonderlund2018LAInterventions}. Much of this empirical tradition has been built on widely reused datasets, with the Open University Learning Analytics Dataset (OULAD) serving as a recurring point of reference across both Learning Analytics and Educational Data Mining \citep{Kuzilek2017OULAD}.

A growing body of work recognizes that dropout risk should not be treated as a static binary outcome but as a process that emerges and evolves across the student trajectory. In online and blended settings, this risk can be observed through sequences of interaction, patterns of engagement, and shifts in academic behaviour, bringing dropout prediction closer to explicitly temporal formulations \citep{Mubarak2020InteractionLogs}. The relevant question then becomes not only which students are at risk, but when that risk intensifies and at what point an intervention might be most useful. Early warning systems built around weekly windows or incremental course progression percentages reflect this shift, treating the timing of prediction as integral to its analytical value \citep{Akcapinar2019EarlyWarning,Adnan2021EarlyIntervention}. More recent work shows that both predictor importance and model performance can change over time from enrolment, reinforcing the view of dropout risk as a longitudinal phenomenon rather than a classification problem \citep{Vaarma2024FinnishHE,Seo2024DropoutRisk}. Correct enforcement of temporal observation boundaries (computing predictors exclusively from events prior to the prediction point, without drawing on post-window data) has been identified as a critical leakage-avoidance requirement for temporal LA models; its consequences for apparent predictive performance under standard evaluation protocols have been documented as a methodological concern that warrants explicit verification in benchmark designs \citep{Kapoor2023Leakage}.

Despite these advances, the comparative literature on dropout prediction still faces significant limitations. Studies vary considerably in target outcomes, data sources, feature types, algorithms, and evaluation metrics, producing a fragmented landscape in which direct comparison across findings is difficult \citep{Sghir2022PLAReview}. Even explicitly comparative studies often operate under non-equivalent conditions (differing simultaneously in problem definition, temporal scope, feature availability, input representation, and experimental design). It therefore remains unclear whether reported performance differences reflect genuine advantages of one model over another or simply artefacts of the evaluation context. Coussement~et~al.'s systematic comparison across algorithms, for instance, is motivated precisely by this heterogeneity \citep{Coussement2020LogitLeaf}. For Learning Analytics, the lack of unification undermines both the methodological cumulativeness of the field and the possibility of translating model findings into dependable guidance for educational practice \citep{Sghir2022PLAReview,Sonderlund2018LAInterventions}.

A related limitation is the evaluation literature's strong centring on aggregate performance measures: accuracy, AUC, F1, or global error metrics \citep{Sghir2022PLAReview,Coussement2020LogitLeaf}. These are useful but insufficient. In Learning Analytics applications, correctly ranking students by risk is not enough; the risk estimates themselves must be coherent enough to support institutional decisions, follow-up prioritization, and resource allocation. Although the field does not always address calibration explicitly, several studies already treat individual risk probabilities as direct input for personalized intervention, making the numerical coherence of those predictions both analytically and practically consequential \citep{Xing2019MOOCDropout,Shiao2023RiskTracking}. More broadly, recent survival-methodology work has emphasized that different metrics capture different aspects of performance and should be interpreted jointly rather than as interchangeable summaries \citep{Vasilev2023SensitivitySurvivalMetrics}. The literature also stresses the importance of interpretability: when institutions need to understand the factors associated with risk and translate them into support strategies, discrimination alone is not enough. Work by Coussement~et~al. and Seo~et~al. illustrates the value of combining strong predictive performance with interpretation that reveals risk factors, segments, and behavioural patterns associated with dropout \citep{Seo2024DropoutRisk}.

A deeper gap lies beneath these methodological concerns. Even when the literature identifies better-performing models, those gains are not consistently translated into an interpretation of what structures predicted dropout risk. Broad reviews of Predictive Learning Analytics continue to emphasize algorithms and performance metrics, with less attention to the substantive question of how model findings connect to risk mechanisms and educational response \citep{Sghir2022PLAReview,Sonderlund2018LAInterventions}. One distinction is particularly consequential: the degree to which performance rests on relatively static student attributes versus temporal and behavioural signals derived from activity across the course. Recent studies confirm that this contrast carries empirical weight \citep{Vaarma2024FinnishHE,Seo2024DropoutRisk}, and it also changes the educational reading of risk, because structural variables and behavioural signals point to different kinds of institutional response. Landmarking approaches and dynamic prediction frameworks extend this idea to continuous-time settings by conditioning predictions on the patient's (or student's) history up to a landmark time, progressively updating survival estimates as longitudinal observations accumulate; the present benchmark adopts a fixed early-window comparable representation as a methodological starting point, with time-varying covariate extensions reserved for future work.

These strands of literature converge on an unresolved need that is methodological as much as empirical: the field lacks sufficiently unified benchmarks that make model comparisons interpretable under a common protocol rather than across heterogeneous settings. The present study addresses that need by treating benchmark design itself as a contribution. Rather than proposing an exhaustive census of all survival architectures, it develops a survival-oriented benchmark structured around two families (Family~A: Dynamic Weekly and Family~B: Static Early-Window with an expanded roster of models), compared under shared evaluation conventions and under an explicit distinction between temporal representation regimes. It then uses ablation, explainability, and calibration to clarify not only which family performs best under the benchmark, but also what kind of information sustains that performance and what that means for Learning Analytics \citep{Sghir2022PLAReview,Coussement2020LogitLeaf}.

Table~\ref{tab:related_work} positions representative studies from this literature against six methodological dimensions that structure the present benchmark. The dimensions are: \textbf{D1}~—~two representational families evaluated as an experimental control; \textbf{D2}~—~survival-specific metrics (IBS, time-dependent concordance) as the primary evaluation protocol; \textbf{D3}~—~OULAD used with weekly or daily VLE temporal granularity; \textbf{D4}~—~benchmark design treated as the primary contribution; \textbf{D5}~—~block-level feature ablation assessed with survival metrics; \textbf{D6}~—~equity or fairness analysis by student subgroup. The table reveals that no prior study satisfies D1 (dual representational families as experimental control) and only one partially addresses D5 (block ablation with survival metrics), which are precisely the dimensions the present benchmark is designed to fill.

\begin{table}[htbp]
\centering
\footnotesize
\caption{Positioning of representative prior studies against six methodological dimensions (D1--D6) of the present benchmark. \textbf{D1}:~dual representation families as experimental control; \textbf{D2}:~survival metrics as primary protocol; \textbf{D3}:~OULAD with weekly/daily VLE granularity; \textbf{D4}:~benchmark design as primary contribution; \textbf{D5}:~block-level ablation with survival metrics; \textbf{D6}:~equity/fairness analysis by subgroup. Symbols: \checkmark~=~present and central; $\sim$~=~partial; --~=~absent.}
\label{tab:related_work}
\begin{tabularx}{\linewidth}{@{}>{\raggedright\arraybackslash}Xcccccc@{}}
\toprule
\textbf{Study} & \textbf{D1} & \textbf{D2} & \textbf{D3} & \textbf{D4} & \textbf{D5} & \textbf{D6} \\
\midrule
Gutierrez-Pachas~et~al.\ (2023) \citep{GutierrezPachas2023Survival}  & --  & \checkmark & --  & $\sim$ & --  & $\sim$ \\
Pan~et~al.\ (2022) \citep{Pan2022SAVSNet}                              & --  & \checkmark & --  & --     & --  & --     \\
Adnan~et~al.\ (2021) \citep{Adnan2021EarlyIntervention}               & --  & --  & $\sim$ & --  & --  & --     \\
He~et~al.\ (2020) \citep{He2020RNNGRU}                                & --  & --  & \checkmark & -- & -- & --     \\
Villar \& de Andrade (2024) \citep{Villar2024Comparative}             & --  & --  & --  & $\sim$ & -- & --     \\
Coussement~et~al.\ (2020) \citep{Coussement2020LogitLeaf}             & --  & --  & --  & $\sim$ & -- & --     \\
Fazil \& Albahlal (2025) \citep{Fazil2025MultiFAR}                    & --  & --  & \checkmark & -- & $\sim$ & -- \\
Gao \& Ni (2025) \citep{Gao2025FairEduNet}                            & --  & --  & $\sim$ & -- & -- & \checkmark \\
Vaarma~et~al.\ (2024) \citep{Vaarma2024FinnishHE}                     & --  & --  & --  & --     & --  & \checkmark \\
Zanellati~et~al.\ (2024) \citep{Zanellati2024Balancing}               & --  & --  & --  & --     & --  & \checkmark \\
\midrule
\textbf{This work}  & \checkmark & \checkmark & \checkmark & \checkmark & \checkmark & $\sim$ \\
\bottomrule
\end{tabularx}
\end{table}

\paragraph*{Study framework.}
Two commitments organize the paper: (i)~dropout is treated as a \emph{temporal-behavioral process}, not a static background trait; and (ii)~consequently, \emph{benchmark design itself} is the contribution---a unified, multi-dimensional protocol organized by representational family---rather than a single best-model claim. Every methodological choice below follows from these two.

\section{Methods}

\begin{figure}[ht]\centering
\includegraphics[width=\linewidth]{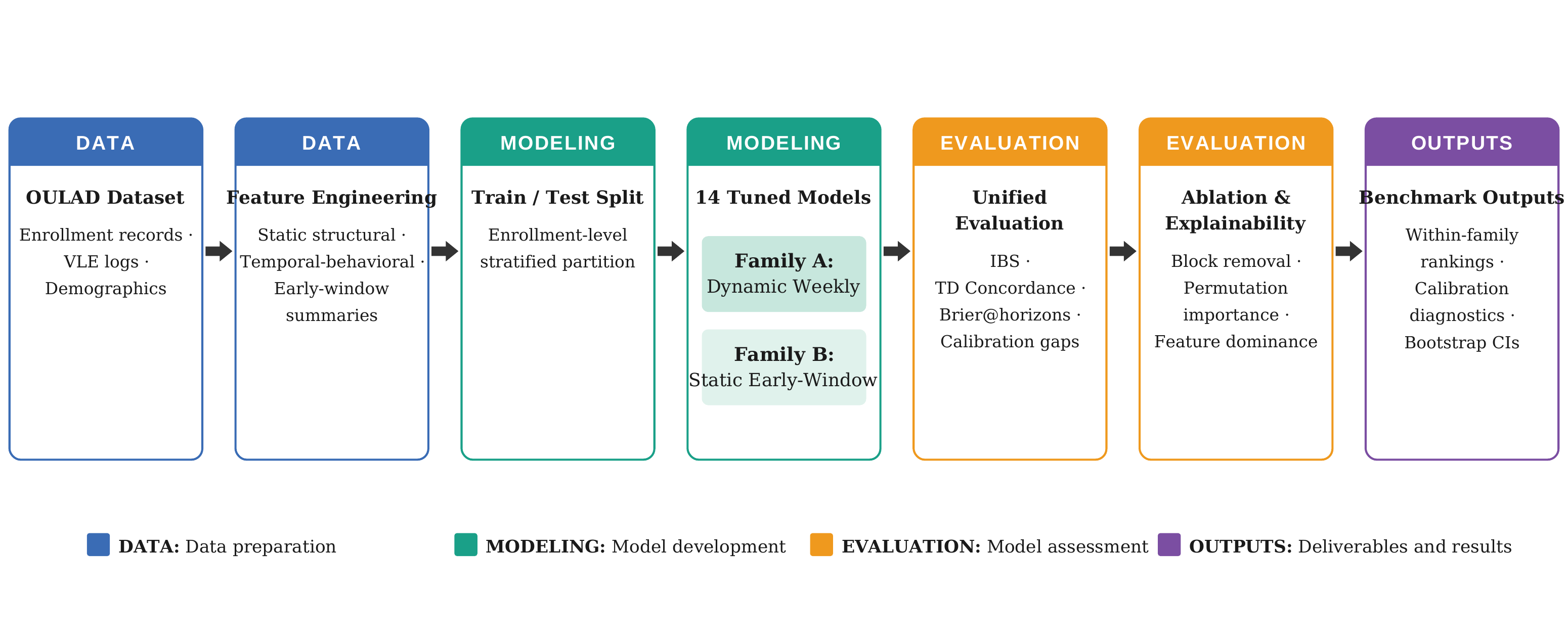}
\caption{Overview of the full benchmark pipeline, organized into four color-coded phases. The \textbf{Data} phase (blue) covers raw data ingestion and feature engineering from OULAD sources into static, temporal-behavioral, and early-window feature blocks. The \textbf{Modeling} phase (teal) encompasses the enrollment-level train/test partition and the training of 14 tuned models across the two representational arms (Family~A: Dynamic Weekly; Family~B: Static Early-Window). The \textbf{Evaluation} phase (orange) applies the unified assessment protocol --- IBS, time-dependent concordance, Brier scores at three horizons, and calibration gaps --- followed by ablation and explainability analyses. The \textbf{Outputs} phase (purple) consolidates within-family rankings, calibration diagnostics, and bootstrap uncertainty estimates as the final benchmark deliverables.}
\label{fig:methods_overview_pipeline}
\end{figure}

To address the unification and interpretability gaps identified above (heterogeneous evaluation protocols, neglected calibration, and insufficient signal attribution), the benchmark applies consistent evaluation protocols across both families while respecting the methodological boundaries between survival model types. Two representational regimes structure the benchmark throughout this paper. \textbf{Family~A: Dynamic Weekly} uses a \textit{person-period} data representation (one row per enrollment per week), in which weekly conditional hazards are estimated and then accumulated into enrollment-level survival curves; this family is also referred to as the weekly discrete-time family. \textbf{Family~B: Static Early-Window} uses a fixed \textit{enrollment-level} representation built from early-window summaries, in which a single feature vector per enrollment feeds directly into continuous-time survival models. These two family labels and their corresponding data representations are used interchangeably throughout the paper and always refer to the same methodological distinction. The two-family design provides a shared evaluation infrastructure (consistent metrics, preprocessing, tuning, and reporting conventions) while respecting the representational distinction between the two families. Figure~\ref{fig:methods_overview_pipeline} provides a structured overview of the full benchmark pipeline. Two orientation tables anchor the rest of the paper: Table~\ref{tab:model_families_overview} situates all 14~models by family, model type, and inductive bias, and Table~\ref{tab:metric_reference} gives a plain-language reading of every metric and when to trust it. Readers can use them as a map before the per-layer detail.

This within-family reporting choice is grounded in the sensitivity of survival metrics to temporal representation. Varying only the discretization or representation scheme shifts the integrated Brier score \citep{Kim2023GNNSurv,KvammeBorgan2019DiscreteSurvival} and produces non-monotonic variation in time-dependent concordance for one and the same model \citep{Nooraldaim2026MultiTaskSurvival}; more broadly, different survival metrics capture different aspects of performance and are not interchangeable \citep{Vasilev2023SensitivitySurvivalMetrics,Park2021ReviewSurvivalPerformance}. Because Family~A and Family~B operate under distinct temporal representations, a single numerical cross-family ranking would conflate genuine model differences with artifacts of temporal representation. Results are therefore reported within each family, over shared enrollment-level horizons, rather than as a cross-family ranking. The cross-family ranking is deliberately \emph{withheld} (not infeasible): it would conflate model differences with representation artifacts, whereas convergent within-family findings are precisely what the two-arm control is designed to expose.

\begin{table}[tbp]
\centering
\footnotesize
\caption{Feature-block membership across the two families. The overlap is \emph{intentional} and confined to the shared \emph{static structural} block; the \emph{exclusive} block is what creates each family's representational regime. Because both families derive from the same OULAD source, agreement across them is not confounded by differing data. Full variable lists are in Table~\ref{tab:feature_blocks}.}
\label{tab:family_overlap}
\begin{tabular}{@{}p{0.30\linewidth}p{0.31\linewidth}p{0.31\linewidth}@{}}
\toprule
Shared (both families) & Family~A only (dynamic weekly) & Family~B only (static early-window) \\
\midrule
\textit{Static structural} block: demographics and prior academic record (OULAD \texttt{studentInfo}). & \textit{Dynamic temporal-behavioral} block: week-by-week engagement signals on the person-period panel. & \textit{Early-window behavioral} block: first-weeks engagement compressed into enrollment-level scalars. \\
\bottomrule
\end{tabular}
\end{table}

\subsection{Dataset}

\begin{sloppypar}
The empirical basis for the benchmark is the \textit{Open University Learning Analytics Dataset} (OULAD), which integrates demographic, academic, assessment, and virtual learning environment (VLE) interaction data \citep{Kuzilek2017OULAD}. The dataset is publicly available at \url{https://research.stem.open.ac.uk/ouanalyse/dataset/}. Direct download at \url{http://schools.stem.open.ac.uk/cdn/files/anonymisedData.zip}. Its continued use across Learning Analytics and Educational Data Mining studies supports its suitability as a benchmark dataset for comparative methodological work \citep{Mihaescu2021PublicDatasets}.
\end{sloppypar}

The dataset links administrative and behavioural information: student demographics, course registration records, assessment data, and large-scale traces of online VLE interaction \citep{Kuzilek2017OULAD}. The central analytical unit is the \textit{enrollment}, defined as the student--module--presentation combination. This unit corresponds to the level at which withdrawal is administratively observed and allows the benchmark to represent both weekly risk trajectories and enrollment-level summaries of early engagement \citep{Prenkaj2020SurveyDropout,Rodriguez2023ChileDropout}.

The benchmark dataset comprises 32,593~enrollments spanning 7~modules and 4~presentations. The observed event rate (enrollments with \textit{Withdrawn} status and a valid unregistration date) is 22.7\%, with the remainder right-censored at the end of their observation window.

\begin{figure}[ht]\centering
\includegraphics[width=.74\linewidth]{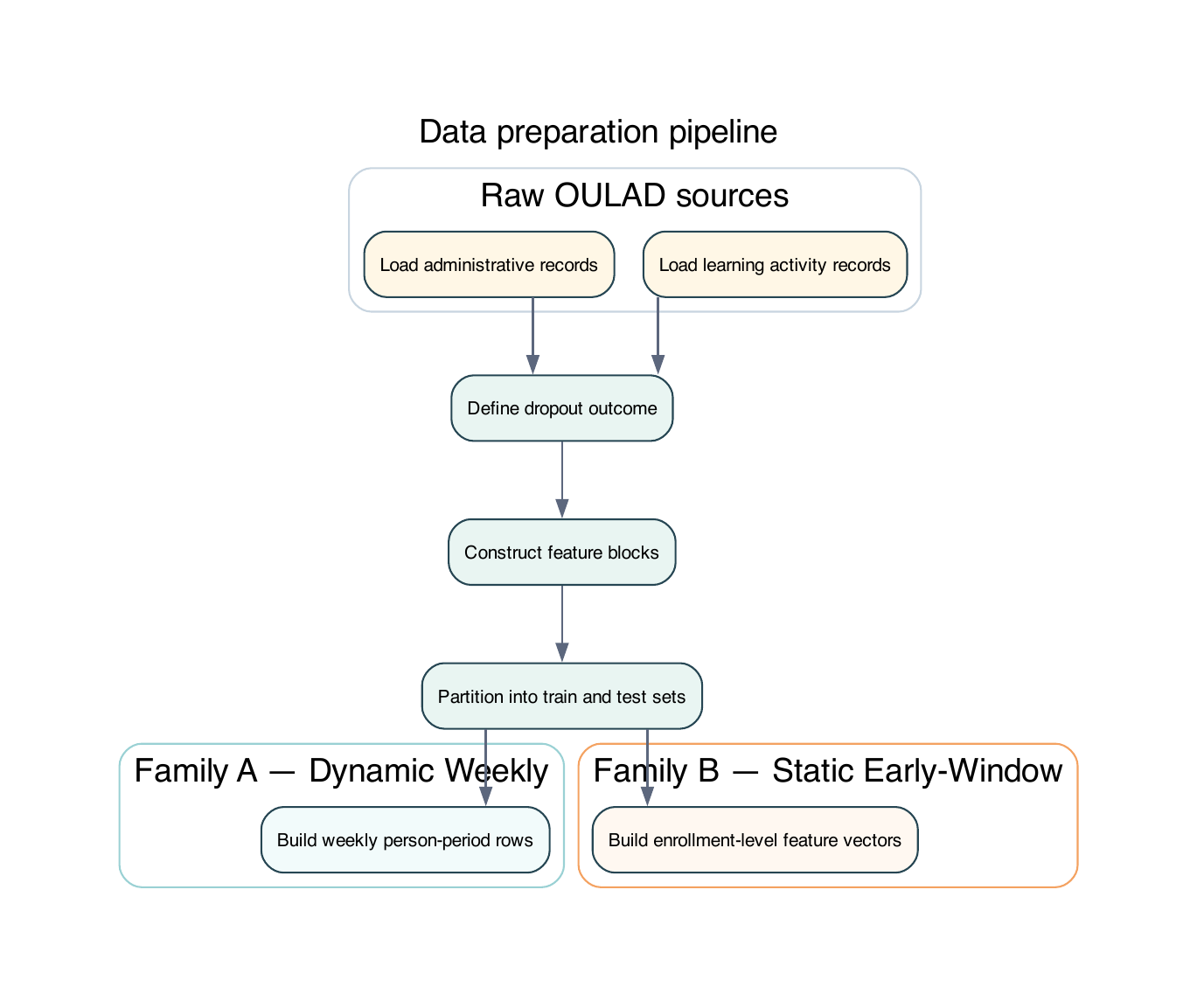}
\caption{Data preparation pipeline from raw OULAD sources to benchmark-ready datasets for the discrete-time and static early-window families.}
\label{fig:methods_data_pipeline}
\end{figure}

\subsection{Outcome definition and prediction task}

The outcome of interest is student dropout, operationalized through the administratively recorded status \textit{Withdrawn} for enrollments with a valid unregistration date. This follows standard practice in dropout modelling, where administrative status transitions, non-reenrolment, or institutional withdrawal records define the target event \citep{Rodriguez2023ChileDropout,Cannistra2021EarlyPredicting}.

For each enrollment $i$, let $T_i$ denote the observed time in weeks and let $\delta_i \in \{0,1\}$ denote the event indicator, where $\delta_i = 1$ indicates an observed withdrawal and $\delta_i = 0$ indicates censoring. Formally,

\begin{equation}
\delta_i =
\begin{cases}
1, & \text{if enrollment } i \text{ has observed withdrawal},\\
0, & \text{otherwise.}
\end{cases}
\label{eq:event_indicator}
\end{equation}

Rather than treating dropout as a static end-of-course binary label, the study formulates the task as \textit{temporal dropout risk prediction}, consistent with work that treats student dropout as a longitudinal or time-dependent process \citep{Prenkaj2020SurveyDropout,Mduma2019SurveyDropout}. Time is discretized at the weekly level, a granularity that is both practically meaningful and widely adopted in educational early-warning settings \citep{Aljohani2019ClickstreamRisk,Adnan2021EarlyIntervention}. Under this design, the benchmark compares models in terms of how they represent temporal dropout risk under common evaluation horizons, not merely whether they predict eventual withdrawal.

\subsection{Data preparation and feature engineering}

Data preparation followed two parallel branches, each designed to accommodate a different temporal representation under a common benchmark framework.

For the discrete-time models, the benchmark adopts a \textit{person-period} representation in which each enrollment is expanded into sequential weekly observations until the event or censoring. This structure follows discrete-time event history modelling, where the conditional hazard is estimated across repeated intervals for each observational unit \citep{Allison1982DiscreteTime,Suresh2022DiscreteTime}. The weekly format supports dynamic temporal features (current activity, recency, streak, and cumulative behavioural summaries), consistent with educational research showing the value of clickstream intensity and evolving activity traces in modelling academic risk \citep{Adnan2021EarlyIntervention,Waheed2020VLEDeepLearning}.

For the continuous-time comparable models, the benchmark uses an enrollment-level representation with \textit{early-window summaries}. Temporal information from the early part of the course is compressed into a fixed vector of aggregated features capturing initial engagement and behavioural participation. This design aligns with survival modelling frameworks that operate on fixed covariates while predicting time-to-event risk \citep{Suresh2022DiscreteTime,Wiegrebe2023DeepSurvivalReview}, and it allows the continuous-time models to remain methodologically comparable to the discrete-time families while preserving a conceptually distinct view of temporal information. This representation does not constitute a general continuous-time formulation with time-dependent covariates; Family~B operates on fixed enrollment-level representations derived from the early observation window. As a consequence, differences in performance between the two families partly reflect this representational constraint and should not be interpreted as evidence about the general superiority of continuous-time over discrete-time survival modelling. Landmarking and joint modelling approaches constitute a natural extension of this design to settings where time-varying covariates are available; these represent an explicit direction for future comparative benchmarking.

\begin{sloppypar}
Features were organized into three conceptually meaningful blocks: (1) static structural covariates, including demographic and prior academic information; (2) dynamic temporal-behavioural signals, derived from week-by-week engagement; and (3) early-window behavioural summaries, used by the static early-window branch. This block-based structure supports the ablation, explainability, and substantive interpretation analyses that follow \citep{Wiegrebe2023DeepSurvivalReview}. Table~\ref{tab:feature_blocks} lists all variables in each block together with their computation method.
\end{sloppypar}

\begin{table}[ht]
\caption{Feature blocks used in the benchmark, with computation method. Static structural covariates are drawn directly from the OULAD \texttt{studentInfo} table. Dynamic temporal-behavioral features are aggregated from \texttt{studentVle} at the enrollment$\times$week level; derived features (\textit{italic}) are computed on the person-period panel. Early-window behavioral summaries collapse the first four weeks of the dynamic panel into enrollment-level scalars. Block membership governs the ablation and grouped permutation importance analyses.}
\label{tab:feature_blocks}
\centering
\scriptsize
\setlength{\tabcolsep}{3pt}
\renewcommand{\arraystretch}{1.18}
\begin{tabularx}{\linewidth}{@{}
  >{\raggedright\arraybackslash}p{3.8cm}
  >{\raggedright\arraybackslash}p{4.5cm}
  >{\raggedright\arraybackslash}X @{}}
\toprule
\textbf{Block / Variable} & \textbf{Description} & \textbf{Computation} \\
\midrule
\multicolumn{3}{l}{\textit{Static structural covariates --- all models}} \\
\addlinespace[2pt]
\quad\texttt{gender} & Student gender & Direct from \texttt{studentInfo} \\
\quad\texttt{region} & UK region of residence & Direct from \texttt{studentInfo} \\
\quad\texttt{highest\_education} & Highest prior qualification & Direct from \texttt{studentInfo} \\
\quad\texttt{imd\_band} & Index of Multiple Deprivation decile & Direct from \texttt{studentInfo} \\
\quad\texttt{age\_band} & Age group at registration & Direct from \texttt{studentInfo} \\
\quad\texttt{disability} & Declared disability status & Direct from \texttt{studentInfo} \\
\quad\texttt{num\_of\_prev\_attempts} & Prior module attempts & Direct from \texttt{studentInfo} \\
\quad\texttt{studied\_credits} & Credits enrolled in module & Direct from \texttt{studentInfo} \\
\addlinespace[5pt]
\multicolumn{3}{l}{\textit{Dynamic temporal-behavioral --- Family~A only}} \\
\addlinespace[2pt]
\quad\texttt{total\_clicks\_week} & VLE clicks in current week & \texttt{SUM(sum\_click)} from \texttt{studentVle}, by enrollment$\times$week \\
\quad\texttt{n\_vle\_rows\_week} & VLE interaction rows this week & \texttt{COUNT(*)} from \texttt{studentVle}, by enrollment$\times$week \\
\quad\texttt{active\_this\_week} & Any VLE activity this week (binary) & 1 if any \texttt{studentVle} row exists for that enrollment$\times$week \\
\quad\texttt{n\_distinct\_sites\_week} & Distinct VLE sites visited this week & \texttt{COUNT(DISTINCT id\_site)} from \texttt{studentVle}, by enrollment$\times$week \\
\quad\textit{\texttt{cum\_clicks\_until\_t}} & Cumulative VLE clicks up to week $t$ & Running \texttt{SUM(total\_clicks\_week)} over expanding window ordered by week \\
\quad\textit{\texttt{recency}} & Weeks since last active week & 0 if active this week; $t - t_{\text{last active}}$ otherwise; $t+1$ if never active \\
\quad\textit{\texttt{streak}} & Consecutive active weeks up to $t$ & Row number within current uninterrupted active run; 0 if inactive \\
\addlinespace[5pt]
\multicolumn{3}{l}{\textit{Early-window behavioral summaries --- Family~B only (weeks 0--3)}} \\
\addlinespace[2pt]
\quad\texttt{clicks\_first\_4\_weeks} & Total VLE clicks in weeks 0--3 & \texttt{SUM(total\_clicks\_week)} for week $< 4$, grouped by enrollment \\
\quad\texttt{active\_weeks\_first\_4} & Active weeks count in weeks 0--3 & \texttt{SUM(active\_this\_week)} for week $< 4$, grouped by enrollment \\
\quad\texttt{mean\_clicks\_first\_4\_weeks} & Mean clicks per active week in weeks 0--3 & \texttt{clicks\_first\_4\_weeks / active\_weeks\_first\_4}; 0 if never active \\
\bottomrule
\end{tabularx}
\renewcommand{\arraystretch}{1.0}
\setlength{\tabcolsep}{6pt}
\end{table}

Preprocessing was standardized across models as far as their input representations allowed. Numeric variables were imputed with the median and categorical variables with an explicit missing category. Categorical predictors were encoded through one-hot expansion with unknown categories ignored at test time, and numeric variables were standardized. Crucially, all preprocessing transformations were fitted on the training partition only and then applied to validation and test data, so that no information from held-out observations influenced feature transformation.

Class imbalance handling was intentionally conservative: no explicit resampling or reweighting was applied across any of the 14 tuned families. A sensitivity experiment confirmed that explicit inverse-frequency class weighting enlarged rather than reduced Family~A calibration gaps (weighted IBS 0.44--0.52, calibration gap 0.24--0.30 at horizon 10, versus unweighted IBS 0.14, gap 0.09--0.10), validating the unweighted design choice.

Family~A uses only person-period information observed up to horizon $t$; Family~B generates a full survival curve from fixed early-window inputs (main window = first four weeks), preventing use of post-window information. A sensitivity grid ($w \in \{2, 4, 6, 8, 10\}$ weeks) was evaluated across window lengths; results are reported at the canonical window ($w=4$ weeks), which was held fixed for all primary benchmark comparisons (Table~\ref{tab:appendix_window_sensitivity}).

See Figure~\ref{fig:methods_data_pipeline} for a schematic summary of the dual data preparation pipeline.

\subsection{Models}

The benchmark compares 14 tuned models organized across Family~A and Family~B. This methodological breadth is deliberate: conclusions about hierarchies or dominant predictive signals are more defensible when they emerge consistently across families with structurally different inductive biases (linear, tree-based, parametric, and neural) than when they rest on a single representative. Family~A: Dynamic Weekly comprises five models operating on the person-period representation: (1) a linear discrete-time hazard model, (2) a neural discrete-time survival model, (3) a Poisson piecewise-exponential model, (4) a gradient-boosted weekly hazard model, and (5) a CatBoost weekly hazard model. Family~B: Static Early-Window comprises nine models operating on fixed enrollment-level early-window representations: (1) a comparable Cox model, (2) DeepSurv, (3) Random Survival Forest, (4) Gradient-Boosted Cox, (5) Weibull AFT, (6) Royston-Parmar, (7) XGBoost AFT, (8) Neural-MTLR, and (9) DeepHit. Table~\ref{tab:model_families_overview} provides a structured overview of all 14 families organized by representational group, model type, and inductive bias. The equations that follow describe the mathematical anchors of each family; all other families share the same input representation as their respective family anchor.

\begin{table}[ht]
\caption{Overview of the 14~models organized by representational group (Family~A/B), model type, and inductive bias. Models within each group are ordered from linear to neural, with tree-based and parametric families in between.}
\centering
\scriptsize
\setlength{\tabcolsep}{4pt}
\renewcommand{\arraystretch}{1.15}
\begin{tabularx}{\linewidth}{@{}>{ \raggedright\arraybackslash}p{1.8cm}
                              >{\raggedright\arraybackslash}p{2.4cm}
                              >{\raggedright\arraybackslash}X
                              >{\raggedright\arraybackslash}p{3.2cm}@{}}
\toprule
\textbf{Group} & \textbf{Model type} & \textbf{Model} & \textbf{Inductive bias} \\
\midrule
\multirow{5}{*}{\textit{Family~A}} 
  & \multirow{2}{*}{Linear}
    & Linear Discrete-Time Hazard & Linear \\
  & & Poisson Piecewise-Exponential & Linear (log-rate) \\
  & Neural & Neural Discrete-Time Survival & Neural \\
  & \multirow{2}{*}{Gradient-Boosted}
    & GB Weekly Hazard & Tree ensemble \\
  & & CatBoost Weekly Hazard & Tree ensemble \\
\midrule
\multirow{9}{*}{\textit{Family~B}}
  & Cox & Cox Comparable & Linear (PH) \\
  & Neural Cox & DeepSurv & Neural (PH) \\
  & \multirow{2}{*}{Parametric}
    & Weibull AFT & Parametric (AFT) \\
  & & Royston-Parmar & Parametric (spline PH) \\
  & \multirow{3}{*}{Tree ensemble}
    & Random Survival Forest & Tree ensemble \\
  & & Gradient-Boosted Cox & Tree ensemble (PH) \\
  & & XGBoost AFT & Tree ensemble (AFT) \\
  & \multirow{2}{*}{Neural}
    & Neural-MTLR & Neural \\
  & & DeepHit & Neural \\
\bottomrule
\end{tabularx}
\renewcommand{\arraystretch}{1.0}
\setlength{\tabcolsep}{6pt}
\vspace{4pt}
\begin{minipage}{\linewidth}
\scriptsize
\textit{Note.} \textbf{Family~A: Dynamic Weekly} $\equiv$ person-period representation (one row per enrollment per week; weekly hazards accumulated into enrollment-level survival curves). \textbf{Family~B: Static Early-Window} $\equiv$ fixed enrollment-level representation (one row per enrollment; early-window summaries fed directly into continuous-time survival models). The two family labels and their corresponding data representations are used interchangeably throughout the paper.
\end{minipage}
\label{tab:model_families_overview}
\end{table}

The first anchor family is a \textit{linear discrete-time hazard model}, implemented on the person-period dataset. For enrollment $i$ at week $t$, the discrete-time hazard is defined as

\begin{equation}
h_i(t) = P(T_i = t \mid T_i \ge t, \mathbf{x}_{it}),
\label{eq:discrete_hazard}
\end{equation}

where $\mathbf{x}_{it}$ denotes the covariates observed for enrollment $i$ at week $t$. In the linear specification, the weekly conditional hazard is linked to the predictors through

\begin{equation}
\text{logit}\big(h_i(t)\big) = \alpha_t + \mathbf{x}_{it}^{\top}\beta,
\label{eq:linear_discrete_hazard}
\end{equation}

where $\alpha_t$ is a time-specific intercept and $\beta$ is the coefficient vector \citep{Suresh2022DiscreteTime}.

The enrollment-level survival curve induced by the weekly hazards is then reconstructed as

\begin{equation}
S_i(t) = P(T_i > t \mid \mathbf{x}_{i1}, \ldots, \mathbf{x}_{it}) = \prod_{\tau = 1}^{t}\big(1 - h_i(\tau)\big),
\label{eq:discrete_survival_reconstruction}
\end{equation}

with fixed-horizon event risk obtained by complementarity,

\begin{equation}
r_i(t) = P(T_i \le t \mid \mathbf{x}_{i1}, \ldots, \mathbf{x}_{it}) = 1 - S_i(t).
\label{eq:discrete_risk_from_survival}
\end{equation}

The Family~A's neural anchor is a \textit{neural discrete-time survival model}, also implemented on the person-period representation. Instead of assuming a linear relationship between covariates and the interval hazard, this family uses a neural architecture to model the weekly hazard flexibly across intervals. Conceptually, this replaces the linear predictor in Equation~\ref{eq:linear_discrete_hazard} with a nonlinear function $f_{\theta}(\mathbf{x}_{it})$, allowing more complex interactions while preserving interval-based time-to-event prediction \citep{KvammeBorgan2019DiscreteSurvival,Wiegrebe2023DeepSurvivalReview}.

\begin{equation}
h_i(t) = \sigma\!\left(\alpha_t + f_{\theta}(\mathbf{x}_{it})\right),
\label{eq:neural_discrete_hazard}
\end{equation}

where $\sigma(z) = \big(1 + e^{-z}\big)^{-1}$ is the logistic link.

The Family~B's linear anchor is a \textit{comparable Cox model}, estimated at the enrollment level using the early-window summaries. The Cox proportional hazards model specifies the hazard as

\begin{equation}
\lambda_i(t \mid \mathbf{x}_i) = \lambda_0(t)\exp(\mathbf{x}_i^\top\beta),
\label{eq:cox_model}
\end{equation}

where $\lambda_0(t)$ is the baseline hazard and $\mathbf{x}_i$ is the fixed covariate vector for enrollment $i$ \citep{Cox1972RegressionModels}.

The Family~B's neural anchor is \textit{DeepSurv}, also estimated at the enrollment level. DeepSurv extends the Cox model by replacing the linear predictor with a neural network, yielding

\begin{equation}
\lambda_i(t \mid \mathbf{x}_i) = \lambda_0(t)\exp\big(f_{\theta}(\mathbf{x}_i)\big),
\label{eq:deepsurv_model}
\end{equation}

where $f_{\theta}(\mathbf{x}_i)$ is the nonlinear representation learned by the network \citep{Katzman2016DeepSurv,Kvamme2019TimeToEvent}.

The primary comparative focus is on tuned versions, in which models are compared under systematically optimized rather than default parameterizations \citep{Kvamme2019TimeToEvent}.

\subsection{Unified evaluation protocol}

All tuned models were assessed under a unified survival-oriented evaluation protocol designed to reduce interpretive ambiguity across differing problem definitions, temporal scopes, and metric selections \citep{Herrmann2021SurvivalBenchmark}.

Evaluation was organized around a hierarchy of complementary survival-oriented metrics. The main metrics include the \textit{Integrated Brier Score} (IBS), \textit{time-dependent concordance}, horizon-specific Brier scores, and horizon-specific calibration at the shared benchmark horizons of weeks 10, 20, and 30. For a horizon $t$, let $\hat{G}(t)$ denote the Kaplan--Meier estimate of the censoring survival function. The IPCW Brier score used in the benchmark may be expressed as

\begin{equation}
\text{BS}^{\mathrm{IPCW}}(t) = \frac{1}{n}\sum_{i=1}^{n}\left[
\frac{\mathbb{I}(T_i \le t,\, \delta_i = 1)}{\hat{G}(T_i^-)}\hat{S}_i(t)^2
\;+
\frac{\mathbb{I}(T_i > t)}{\hat{G}(t)}\big(1 - \hat{S}_i(t)\big)^2
\right],
\label{eq:brier_score}
\end{equation}

where $\hat{S}_i(t)$ is the predicted survival probability for enrollment $i$ at horizon $t$. The integrated version reported throughout the article is

\begin{equation}
\mathrm{IBS} = \frac{1}{\tau_{\max}}\int_{0}^{\tau_{\max}} \text{BS}^{\mathrm{IPCW}}(u)\,du,
\label{eq:ibs}
\end{equation}

where $\tau_{\max}$ is the upper evaluation horizon. In the empirical implementation, horizon-specific Brier scores and IBS were computed under right-censoring through inverse-probability-of-censoring weighting (IPCW) using a Kaplan--Meier estimate of the censoring distribution, following the evaluation convention implemented through \texttt{pycox}. IBS summarizes prediction error over the evaluation interval and served as the main global measure of probabilistic performance \citep{Graf1999PrognosticSchemes,Gerds2006Brier,Park2021ReviewSurvivalPerformance}.

Discrimination was evaluated through \textit{time-dependent concordance} in the formulation of Antolini~et~al.\ \citep{Antolini2005TimeDependentDiscrimination}, which assesses concordance of predicted survival functions with observed event ordering across the full event-time horizon, rather than through an unspecified static-risk C-index. This distinction matters because the benchmark compares full survival trajectories across common horizons, not merely fixed scalar risk scores. Accordingly, this article refers to this quantity as a time-dependent concordance measure, even when compact tables retain the conventional shorthand \textit{C-index} \citep{Harrell1982YieldMedicalTests,Longato2020CIndex,Park2021ReviewSurvivalPerformance}.

The protocol distinguishes dynamic from static early-window prediction settings: discrete-time families update weekly hazards sequentially using only week-$t$ information, accumulating enrollment-level survival as in Equations~\ref{eq:discrete_survival_reconstruction}--\ref{eq:discrete_risk_from_survival}; static early-window families generate a full survival curve from fixed early-window inputs. Both branches report at shared enrollment-level horizons without information leakage.

The enrollment-level split was stratified by event status with no identity leakage; a contextual audit confirmed that curricular context (all 7~modules, 4~presentations, 22 module-presentation combinations) was fully shared across train and test. Findings generalize across enrollments under shared curricular context, not to unseen module or presentation settings.

Horizon-specific calibration was evaluated through a bin-based risk calibration procedure. At each benchmark horizon $h$, predicted event risk was grouped into quantile-based bins $b = 1, \ldots, B$, and calibration error was summarized as the sample-size-weighted mean absolute gap

\begin{equation}
\mathrm{Calib}(h) = \sum_{b=1}^{B}\frac{n_b}{n}\left|\bar{r}_b(h) - \tilde{y}_b(h)\right|,
\label{eq:calibration_gap}
\end{equation}

where $n_b$ is the size of bin $b$, $\bar{r}_b(h)$ is the mean predicted event risk in that bin, and $\tilde{y}_b(h)$ is the IPCW-adjusted empirical event rate at horizon $h$. The empirical event rates used in this comparison were therefore estimated under right-censoring adjustment, consistent with the IPCW framework applied to the Brier score and IBS computations, rather than as unadjusted observed proportions. These reported calibration values are therefore horizon-specific weighted absolute calibration gaps. In addition, the strengthened calibration audit estimated approximate calibration intercept and slope by horizon through a weighted fit on the reliability-bin summaries, so that calibration could be interpreted jointly through primary gap, supporting Brier, support, and logit-scale slope/intercept diagnostics rather than through a single scalar alone \citep{Park2021ReviewSurvivalPerformance}.

\begin{sloppypar}
Because leading margins were numerically small, uncertainty was assessed through enrollment-level bootstrap resampling (200~resamples, no model refit) on the held-out test set. Frozen predictions were resampled to obtain empirical intervals for IBS, time-dependent concordance, and Brier horizons; these reflect sampling variability only and are lower bounds on total uncertainty; retraining-based intervals, which would propagate model estimation variance, could be substantially wider. For time-dependent concordance, RSF's leading position was preserved in a large majority of resamples; for IBS, the bootstrap provided a directional signal only.
\end{sloppypar}

The comparable Cox model was subjected to a formal proportional-hazards audit yielding a broad-departure-from-proportionality classification (10 of 38~covariates, 26.3\%); results from the comparable Cox branch should be narrated as from an approximate specification. DeepSurv shares the Cox-type structure but is not amenable to classical Schoenfeld diagnostics; detailed scope boundaries are reported in the Appendix.

Table~\ref{tab:metric_reference} summarises all evaluation metrics used in the benchmark, organised by analytical role, with a plain-language description to support readers less familiar with survival-specific evaluation conventions.

\begin{table}[ht]
\caption{Reference summary of all evaluation metrics used in the benchmark. IBS is the survival analogue of the Mean Squared Error (MSE) used in classical ML regression: where MSE averages squared prediction errors on a fixed outcome, IBS averages squared errors on the predicted survival probability over the full event-time axis, adjusted for censoring via IPCW. For error-based metrics, lower scores correspond to stronger performance, whereas discrimination metrics improve as scores increase. Calibration gaps nearer zero reflect closer agreement between predicted risk and observed dropout rates.}
\centering
\scriptsize
\setlength{\tabcolsep}{3pt}
\renewcommand{\arraystretch}{1.18}
\begin{tabularx}{\linewidth}{@{} >{\raggedright\arraybackslash}p{2.6cm}
                                    >{\raggedright\arraybackslash}p{2.0cm}
                                    >{\raggedright\arraybackslash}p{1.5cm}
                                    >{\raggedright\arraybackslash}X @{}}
\toprule
\textbf{Metric} & \textbf{Role} & \textbf{Better when} & \textbf{What it measures} \\
\midrule
\multicolumn{4}{l}{\textit{Primary performance metrics}} \\
\addlinespace[2pt]
IPCW weight $\hat{G}(t)^{-1}$ & Censoring correction & --- & Inverse-probability-of-censoring weight, estimated non-parametrically via Kaplan-Meier on the censoring process. Upweights uncensored observations to correct for the selective removal of censored students from the risk set as follow-up progresses. Applied internally to Brier@h, IBS, and Calib@h; not a standalone evaluation metric. \\
\addlinespace[3pt]
Brier@10, @20, @30 & Error & Lower $\downarrow$ & Horizon-specific probabilistic error: squared difference between predicted survival probability and observed event status at a fixed week, averaged over enrollments with IPCW adjustment. Analogous to MSE at a snapshot in time. \\
\addlinespace[3pt]
Integrated Brier Score (IBS) & Error & Lower $\downarrow$ & Global probabilistic error integrated over the full evaluation window. Survival analogue of MSE: where MSE averages squared errors on a fixed outcome, IBS averages squared errors on the predicted survival curve over time, adjusted for censoring. \\
\addlinespace[3pt]
TD Concordance (C-index) & Discrimination & Higher $\uparrow$ & Time-dependent concordance (Antolini): probability that a student who drops out earlier receives a higher predicted risk than one who drops out later or does not drop out. Measures ranking quality across the full event-time trajectory, not just at a fixed horizon. \\
\addlinespace[5pt]
\multicolumn{4}{l}{\textit{Calibration metrics}} \\
\addlinespace[2pt]
Calib@10, @20, @30 & Calibration & Lower $\downarrow$ & Horizon-specific weighted absolute gap between mean predicted event risk and IPCW-adjusted observed dropout rate within quantile bins. Zero means predicted probabilities match observed rates exactly at that horizon. \\
\addlinespace[3pt]
Calibration slope \& intercept & Calibration (auxiliary) & Slope $\approx 1$, intercept $\approx 0$ & Logit-scale linear fit on reliability-bin summaries. Slope $< 1$ indicates overconfidence (predictions too extreme); slope $> 1$ indicates underconfidence; intercept $\neq 0$ indicates systematic over- or underestimation. Treated as strengthening diagnostics, not primary ranking criteria. \\
\addlinespace[5pt]
\multicolumn{4}{l}{\textit{Ranking stability diagnostic}} \\
\addlinespace[2pt]
Bootstrap rank-1 share / 95\% CI & Stability & Higher share = more stable & Proportion of 200 enrollment-level resamples in which a model holds rank~1 on a given metric (frozen predictions, no model refit). Captures sampling variability in ranking, not estimation variance. Intervals are lower bounds on total uncertainty. \\
\bottomrule
\end{tabularx}
\renewcommand{\arraystretch}{1.0}
\setlength{\tabcolsep}{6pt}
\label{tab:metric_reference}
\end{table}

This multi-metric design reflects the recognition that no single metric can adequately summarize model quality in temporal risk prediction \citep{Park2021ReviewSurvivalPerformance}.

\begin{figure}[ht]\centering
\includegraphics[width=.74\linewidth]{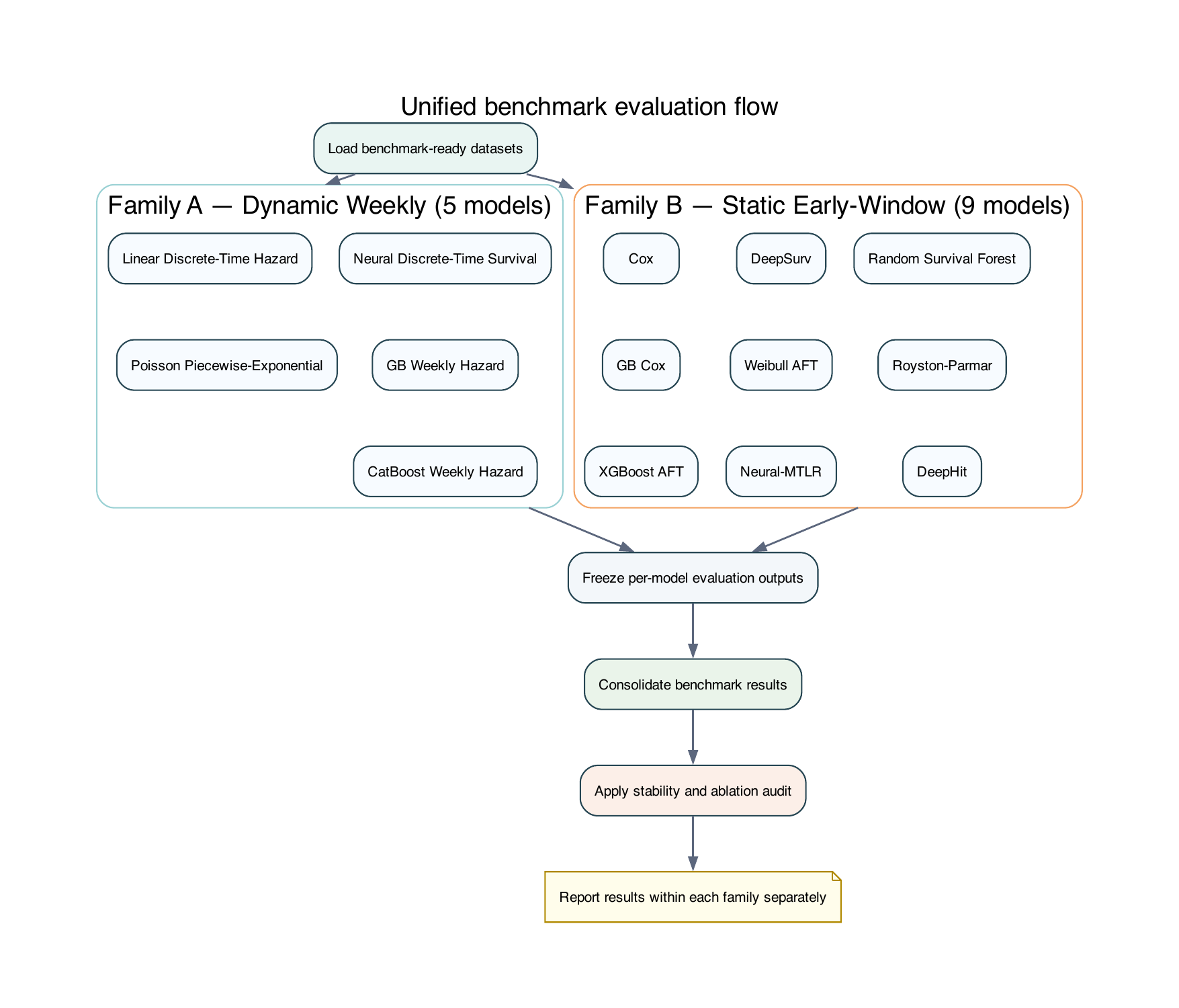}
\caption{Unified benchmark evaluation flow used to compare the models under a common survival-oriented protocol.}
\label{fig:methods_benchmark_flow}
\end{figure}

See Figure~\ref{fig:methods_benchmark_flow} for a schematic overview of the unified benchmark evaluation flow. Additional protocol details, calibration-by-horizon diagnostics, bootstrap uncertainty summaries, and proportional-hazards audit outputs are reported in the Appendix.

\subsection{Ablation design}

The ablation analysis estimates the relative contribution of major feature blocks to model performance. Rather than interpreting performance at the level of the full model alone, the benchmark evaluates tuned variants in which selected predictor groups are removed or isolated, making it possible to assess how far predictive performance depends on static structural covariates versus temporal and behavioural information.

Ablation serves here as both a sensitivity analysis and an interpretive tool. Comparing full-feature models with variants that exclude specific blocks reveals which kinds of signal most strongly sustain predictive performance \citep{LiJanson2024OptimalAblation}.

\subsection{Explainability and calibration overview}

The benchmark includes two complementary interpretive layers: explainability and calibration.

Explainability methods were matched to each model. For intrinsically interpretable models, global interpretation draws on model coefficients and hazard-related effect sizes \citep{Cox1972RegressionModels}. For the nonlinear families, grouped permutation importance provides a model-agnostic measure of the global contribution of individual predictors and conceptually defined feature blocks \citep{Fisher2019AllModelsWrong}. Four blocks structure this analysis: \textit{static structural covariates} (demographic and prior academic background, applicable across all families); \textit{dynamic temporal-behavioral signals} (week-by-week engagement and activity traces, Family~A only); \textit{discrete time index} (the weekly time step as a standalone predictor of conditional hazard, Family~A only; in the linear discrete-time model, the time step is encoded as week-specific intercepts $\alpha_t$ in Equation~\ref{eq:linear_discrete_hazard}, whereas in the neural and gradient-boosted families it enters as an ordinal covariate in $\mathbf{x}_{it}$; the block encompasses both roles as predictors of the weekly conditional hazard); and \textit{early-window behavioral summaries} (compressed early-course engagement aggregates, Family~B only). This block-based organization enables comparison of explanatory patterns across paradigms while anchoring interpretation in the same feature taxonomy used by the ablation layer.

Calibration provides an explicit check on probabilistic coherence. Where discrimination measures whether students are correctly ranked by relative risk, calibration evaluates whether numerical risk estimates remain coherent with observed event frequencies over the benchmark horizons. In the present benchmark, this layer is interpreted through a primary horizon-wise weighted absolute calibration gap together with supporting slope/intercept diagnostics derived from the reliability-bin summaries \citep{Park2021ReviewSurvivalPerformance}.

Because the present study evaluates calibration as a robustness layer rather than as a model-updating study, post-hoc recalibration is treated as a future strengthening step rather than as a procedure applied to the held-out benchmark comparison itself. Reliability diagrams (calibration plots) are included in the Appendix (Figures~\ref{fig:appendix_reliability_comparable_a}--\ref{fig:appendix_reliability_dynamic}) as visual complements to the tabulated calibration gaps; recalibration experiments (isotonic regression, beta calibration, or Platt scaling applied to the frozen held-out predictions) are deferred to future work. The Family~A calibration gaps (0.089--0.098 at horizon 10, increasing to 0.123--0.149 at horizon 30) are the primary candidate for recalibration follow-up, given their systematically larger magnitude relative to the best Family~B models.

\begin{figure}[ht]\centering
\includegraphics[width=.72\linewidth]{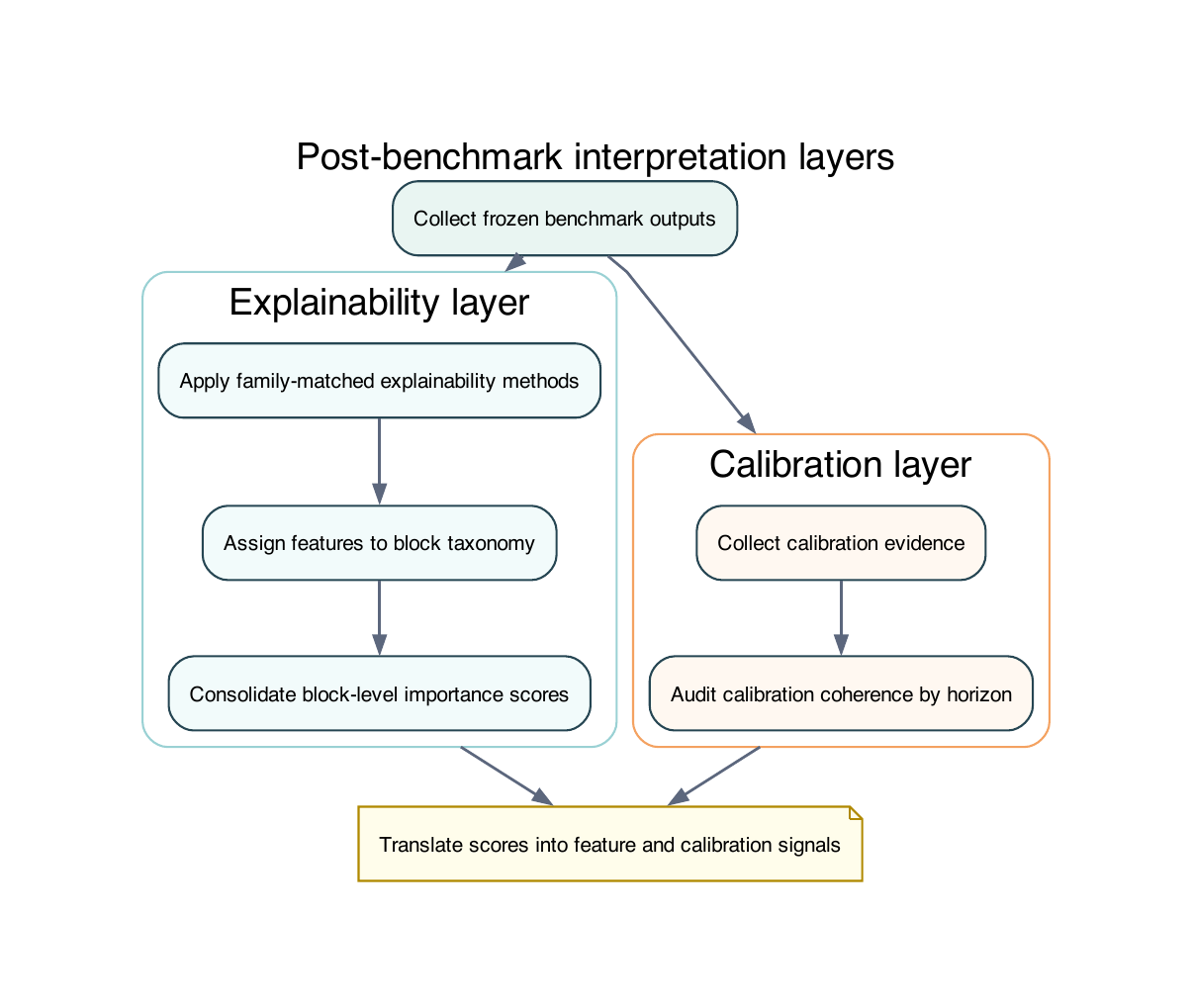}
\caption{Overview of the explainability and calibration layers used as complementary post-benchmark analyses across the models.}
\label{fig:methods_explainability_calibration}
\end{figure}

See Figure~\ref{fig:methods_explainability_calibration} for a schematic overview of how explainability and calibration are integrated into the benchmark.

\subsection{Reproducibility and implementation}

The analytical pipeline was implemented as a reproducible notebook-based workflow, with systematic export of all tables, figures, and metadata reported in the study. This design ensures traceability across data preparation, preprocessing, model fitting, evaluation, ablation, explainability, calibration, and paper integration \citep{Peng2011ReproducibleResearch}. The complete pipeline (including all stage scripts, frozen model artifacts, tables, and figures) is publicly available at \url{https://github.com/rafa-rodriguess/dropout_benchmark}. Every number in the paper is generated by a single export step from these versioned, frozen artifacts, so each reported value traces deterministically to a pipeline output rather than to an intermediate notebook state, a raw database extraction, or an upstream script.

Each of the 14 tuned models followed the same discipline: preprocessing design, enrollment-level validation partitioning, bounded search scope, selection criterion, and complexity controls. Individual records are consolidated in the appendix audit (Table~\ref{tab:appendix_preproc_tuning_audit}).

Tuning was deliberately controlled rather than exhaustive: the goal was stronger, comparable benchmark representatives, not unbounded per-family optimization. All models were subjected to an equivalent tuning budget of 24 candidate evaluations, ensuring that no model benefited from a disproportionately larger search effort. The budget of 24 candidates was set by the most complex model in the benchmark (DeepSurv), whose architecture and regularisation space required the largest grid; all remaining models were tuned under the same ceiling to ensure symmetric search effort.

\subsection*{Use of AI-assisted tools}
During the preparation of this work, the authors used large language model tools (Claude, ChatGPT) to assist with structural revision and editorial drafting of selected sections, with the aim of improving clarity and readability. All outputs were reviewed and edited by the authors, and the authors take full responsibility for the content of this publication. Use was limited to writing assistance; all benchmark design decisions, data analysis, model implementation, and interpretation were performed exclusively by the authors.

\section{Results}

\subsection{Main benchmark comparison}

\paragraph*{Reading guide.}
If you read one line per family: within \textbf{Family~B}, Random Survival Forest leads on time-dependent concordance (0.674) and on all three Brier horizons, with XGBoost AFT as the sole outlier; within \textbf{Family~A}, five models sit inside a 0.0011 IBS band, with Poisson Piecewise-Exponential narrowly ahead. The detail below supports these two lines; the orientation tables are Table~\ref{tab:model_families_overview} (models) and Table~\ref{tab:metric_reference} (metrics).

Results are reported within each family separately. The two families use different risk formulations (weekly person-period hazards versus fixed early-window survival predictions), so that a single combined numerical ranking would conflate genuine model differences with artifacts of temporal representation, to which metrics such as the IBS are known to be sensitive. Within each family, results are organized around the within-family leader---Poisson Piecewise-Exponential for Family~A and Random Survival Forest for Family~B---with the remaining models reported as a comparative roster (Table~\ref{tab:main_benchmark}); the systematic ablation is carried out on a representative eight-model subset, four per family. The full 14-model roster is load-bearing for the main comparison precisely because the within-family spread is itself the result (a 0.0011 IBS band in Family~A; 0.1118--0.1190 for the eight non-outlier models in Family~B): a sparse roster could not establish the full landscape of model performance, while the eight-model subset is the representative cross-section used for signal attribution.

Within Family~B, \textit{Random Survival Forest} showed the highest point estimate for time-dependent concordance (0.674) and the lowest Brier scores across all three horizons; the remaining seven models ranged from 0.1131 to 0.1190 on IBS, with \textit{XGBoost AFT} as the sole outlier (IBS~0.150, TD concordance~0.577). Within Family~A, all five models clustered within a 0.0011 IBS band (0.1399--0.1410), with \textit{Poisson Piecewise-Exponential} showing the lowest point estimate for integrated Brier score.

\begin{table}[hbt]
\caption{Main benchmark comparison across all 14~models, grouped by family. For Family~A, Brier-horizon scores are computed from enrollment-level survival curves reconstructed from accumulated weekly hazards, using the same IPCW methodology as Family~B; both families are evaluated on the full metric set. Lower IBS and Brier scores indicate better probabilistic accuracy (lower mean squared survival error); higher time-dependent concordance indicates better discrimination.}
\centering
\scriptsize
\setlength{\tabcolsep}{4pt}
\begin{tabular}{>{ \raggedright\arraybackslash}p{4.2cm}ccccc}
\toprule
Model & IBS & TD Concordance & Brier@10 & Brier@20 & Brier@30 \\
\midrule
\multicolumn{6}{l}{\textit{Family~A: Dynamic Weekly}} \\
\addlinespace[2pt]
Poisson Piecewise-Exponential & 0.1399 & 0.5019 & 0.1213 & 0.1621 & 0.1931 \\
Linear Discrete-Time Hazard & 0.1400 & 0.5013 & 0.1195 & 0.1637 & 0.1971 \\
CatBoost Weekly Hazard & 0.1408 & 0.5011 & 0.1203 & 0.1644 & 0.1980 \\
Neural Discrete-Time Survival & 0.1410 & 0.4992 & 0.1199 & 0.1648 & 0.1990 \\
Gradient-Boosted Weekly Hazard & 0.1410 & 0.5018 & 0.1206 & 0.1646 & 0.1981 \\
\addlinespace[4pt]
\multicolumn{6}{l}{\textit{Family~B: Static Early-Window}} \\
\addlinespace[2pt]
Random Survival Forest & 0.1118 & 0.6735 & 0.0947 & 0.1329 & 0.1574 \\
Neural-MTLR & 0.1131 & 0.6569 & 0.0954 & 0.1340 & 0.1588 \\
DeepSurv & 0.1145 & 0.6619 & 0.0978 & 0.1343 & 0.1589 \\
Gradient-Boosted Cox & 0.1148 & 0.6621 & 0.1000 & 0.1361 & 0.1601 \\
DeepHit & 0.1151 & 0.5835 & 0.0973 & 0.1363 & 0.1614 \\
Cox Comparable & 0.1176 & 0.6512 & 0.1012 & 0.1378 & 0.1623 \\
Weibull AFT & 0.1180 & 0.6131 & 0.1016 & 0.1380 & 0.1624 \\
Royston-Parmar & 0.1190 & 0.6430 & 0.1035 & 0.1398 & 0.1641 \\
XGBoost AFT & 0.1495 & 0.5772 & 0.1241 & 0.1773 & 0.2174 \\
\bottomrule
\end{tabular}
\setlength{\tabcolsep}{6pt}
\label{tab:main_benchmark}
\end{table}

\begin{figure}[ht]\centering
\includegraphics[width=\linewidth]{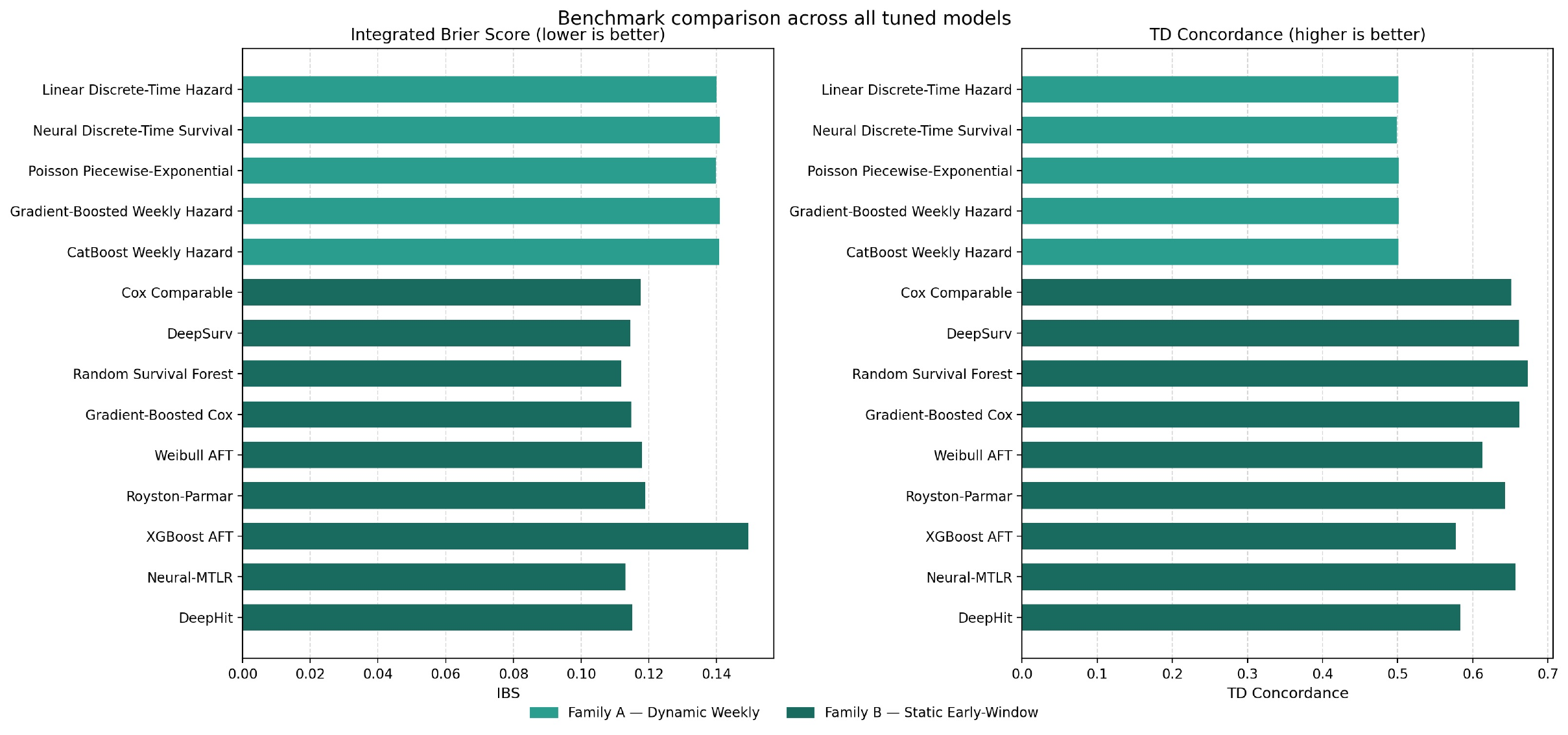}
\caption{Benchmark comparison of all 14~models under the unified survival-oriented protocol, grouped by family. Within Family~B, RSF leads on time-dependent concordance and all Brier horizons; XGBoost AFT is a visible outlier. Within Family~A, all five models cluster tightly on IBS and horizon-specific Brier; Brier values for Family~A are reconstructed from accumulated weekly hazards.}
\label{fig:benchmark_tuned_comparison}
\end{figure}

Within Family~B, the RSF--DeepSurv gap on TD concordance was 0.012; the Brier@10 gap between RSF and the next-best model was 0.001. RSF showed the best point estimate on all reported metrics within that family; these differences may not be statistically significant, as the no-refit bootstrap reflects sampling variability only. The no-refit bootstrap showed RSF leading time-dependent concordance in the majority of resamples and maintaining Brier-horizon leadership under resampling; IBS differences among the top eight Family~B models were modest, with RSF leading at 0.1118 and the next-best cluster ranging from 0.1131 to 0.1190. Within Family~A, \textit{Linear Discrete-Time Hazard} showed the lowest Brier@10 (0.1195) and \textit{Poisson Piecewise-Exponential} showed the lowest Brier@20 and Brier@30 (0.1621 and 0.1931); all five models fell within a 0.0011 IBS band and a 0.0018 Brier@10 band, with no single model dominating across all metrics simultaneously.

The proportional-hazards audit on the comparable Cox model assigned a global classification of broad departure from proportionality: 10 of 38 tested covariates (26.3\%) showed evidence of possible non-proportionality at $\alpha = 0.05$. No other Family~B model was subjected to an identical classical proportional-hazards diagnostic.

The train--test partition was leakage-free at the enrollment identity level; a contextual audit confirmed complete overlap of modules, presentations, and module-presentation combinations across splits.

\subsection{Ablation results: static versus temporal-behavioral signal}

A systematic ablation experiment was conducted on eight of the fourteen tuned models: four from Family~A (\textit{Poisson Piecewise-Exponential}, \textit{Linear Discrete-Time Hazard}, \textit{GB Weekly Hazard}, and \textit{Neural Discrete-Time Survival}) and four from Family~B (\textit{Cox Comparable}, \textit{DeepSurv}, \textit{RSF}, and \textit{Neural-MTLR}), removing each feature block in turn while holding all other modelling choices fixed. The eight models span linear, gradient-boosted tree, and neural architectures across both families, ensuring that the finding on temporal signal dominance does not rest on a single modelling paradigm or a single model. For Family~B, IBS-based and concordance-based ablation are both reported: ablation is performed at the enrollment level under the same fixed early-window representation used for the primary benchmark, so each feature-removal variant yields a well-defined enrollment-level survival curve that can be scored with IBS under the IPCW estimator. For Family~A, only concordance-based ablation is reported: model predictions are person-period hazard scores evaluated row-by-row across weekly intervals, whereas IBS is defined over enrollment-level survival curves reconstructed from those hazards; computing IBS under each feature-removal variant would require re-running the full survival-curve reconstruction pipeline for every ablated model, which was outside the scope of the current benchmark freeze; concordance-based ablation captures the ranking impact of feature removal under the family's native discrimination objective and is methodologically consistent with the family's evaluation contract.

Within Family~B, all four models showed a larger IBS increase when the early-window behavioral block (compressed early-course engagement aggregates, including clickstream activity counts and session summaries; see Explainability layer in Methods) was removed than when static covariates (demographic and prior academic background variables) were removed; on average, temporal signal removal increased IBS by 0.0083 versus 0.0051 for static removal, with IBS ratios ranging from 1.20 (RSF) to 2.46 (Cox Comparable). Across all eight models, concordance decreased more when temporal signal was withheld than when static covariates were removed, with Family~A temporal removal reducing concordance by 0.081--0.102 and static removal causing smaller changes across all four Family~A models.

\begin{table}[hbt]
\caption{Ablation results: Family~B (early-window continuous-time survival). IBS ratios above 1.0 indicate greater dependence on temporal-behavioral signal than on static covariates.}
\centering
\scriptsize
\setlength{\tabcolsep}{4pt}
\begin{tabular}{>{ \raggedright\arraybackslash}p{3.8cm}ccccc}
\toprule
Model & $\Delta$IBS static & $\Delta$IBS temporal & $\Delta$TD static & $\Delta$TD temporal & IBS ratio \\
\midrule
Cox Comparable & 0.0020 & 0.0049 & $-$0.0288 & $-$0.0598 & 2.46 \\
DeepSurv & 0.0044 & 0.0082 & $-$0.0361 & $-$0.0711 & 1.86 \\
Neural-MTLR & 0.0050 & 0.0095 & $-$0.0385 & $-$0.0747 & 1.90 \\
Random Survival Forest & 0.0089 & 0.0106 & $-$0.0510 & $-$0.0764 & 1.20 \\
\bottomrule
\end{tabular}
\setlength{\tabcolsep}{6pt}
\label{tab:ablation_comparable}
\end{table}

\begin{table}[hbt]
\caption{Ablation results: Family~A (weekly discrete-time hazard). Only concordance-based evidence is reported ($\Delta$TD concordance). IBS-based ablation was not computed for Family~A models: the per-person-period hazard evaluation surface differs in temporal footprint from the enrollment-level survival curve on which IBS is defined, and converting the ablation to IBS would require re-running the full curve-reconstruction pipeline under each feature-removal variant. Concordance-based ablation is therefore the methodologically consistent metric for this family.}
\centering
\scriptsize
\setlength{\tabcolsep}{4pt}
\begin{tabular}{>{ \raggedright\arraybackslash}p{4.2cm}cc}
\toprule
Model & $\Delta$TD static & $\Delta$TD temporal \\
\midrule
Gradient-Boosted Weekly Hazard & $-$0.0513 & $-$0.1023 \\
Linear Discrete-Time Hazard & $-$0.0434 & $-$0.0976 \\
Neural Discrete-Time Survival & $-$0.0178 & $-$0.0812 \\
Poisson Piecewise-Exponential & $-$0.0389 & $-$0.0997 \\
\bottomrule
\end{tabular}
\setlength{\tabcolsep}{6pt}
\label{tab:ablation_dynamic}
\end{table}

\begin{figure}[ht]\centering
\includegraphics[width=\linewidth]{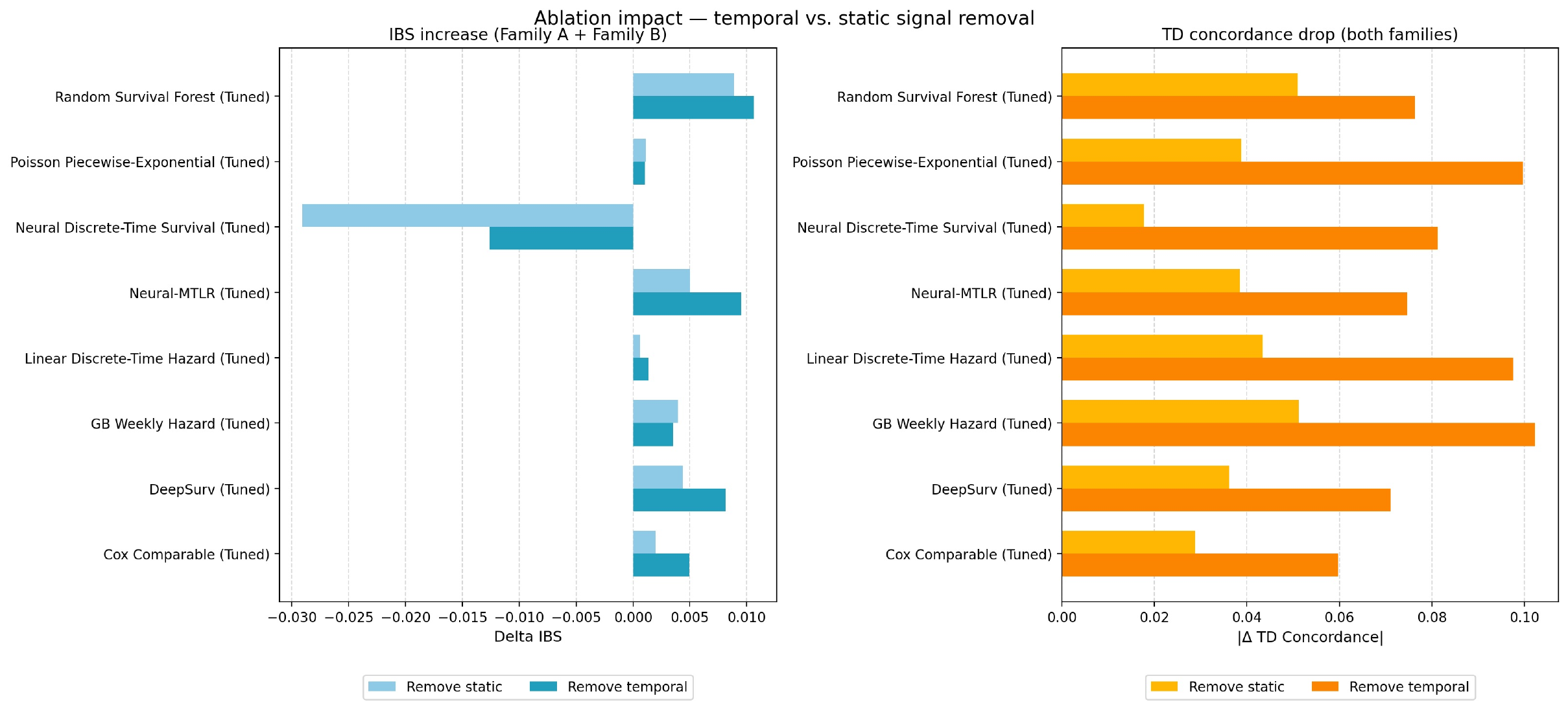}
\caption{Ablation impact across eight models from both families, contrasting the loss associated with removing static versus temporal-behavioral signal. Across all families, removing temporal-behavioral signal is more damaging than removing static covariates. The IBS panel (left) shows only the four Family~B models because IBS-based ablation requires enrollment-level survival curves, which are directly available for the early-window Family~B but not for Family~A, where predictions are weekly hazard scores. The TD concordance panel (right) includes all eight models.}
\label{fig:ablation_impact}
\end{figure}

\subsection{Explainability across models}

The paper-facing explainability export covers eight tuned models spanning both families: four from Family~A (\textit{Poisson Piecewise-Exponential}, \textit{Linear Discrete-Time Hazard}, \textit{GB Weekly Hazard}, and \textit{Neural Discrete-Time Survival}) and four from Family~B (\textit{Cox Comparable}, \textit{DeepSurv}, \textit{RSF}, and \textit{Neural-MTLR}). These models provide complementary inference across linear, gradient-boosted tree, and neural architectures in both the weekly person-period and early-window enrollment-level representations. In all eight models, the dominant feature \textit{block} is temporal: Family~A models show either \textit{dynamic\_temporal\_behavioral} or \textit{discrete\_time\_index} as the dominant block; Family~B models uniformly show \textit{early\_window\_behavior} as dominant. The top individual driver is a temporal or early-window variable in seven of the eight models; in \textit{GB Weekly Hazard}, the highest individual-feature attribution corresponds to a static covariate (\textit{studied\_credits}), yet the dominant block at the aggregate level remains \textit{discrete\_time\_index}, consistent with the temporal dominance pattern across the full ensemble. This distinction between individual-feature attribution and block-level dominance reflects the dual role of the \textit{discrete\_time\_index} block, which encompasses both time-specific intercepts and ordinal time-step covariates depending on the model.

\begin{table}[hbt]
\caption{Cross-family explainability summary for eight models across both families. \textit{Early-window behavioral}: compressed early-course engagement summaries. \textit{Dynamic temporal-behavioral}: week-by-week activity features. \textit{Discrete time index}: the time-step covariate (week number).}
\centering
\scriptsize
\setlength{\tabcolsep}{2pt}
\begin{tabular}{>{ \raggedright\arraybackslash}p{3.8cm}>{ \raggedright\arraybackslash}p{3.4cm}>{ \raggedright\arraybackslash}p{3.6cm}}
\toprule
Model & Top driver & Dominant block \\
\midrule
\multicolumn{3}{l}{\textit{Family~B}} \\
\addlinespace[2pt]
Cox Comparable & clicks\_\allowbreak{}first\_\allowbreak{}4\_\allowbreak{}weeks & early\_\allowbreak{}window\_\allowbreak{}behavior \\
DeepSurv & active\_\allowbreak{}weeks\_\allowbreak{}first\_4 & early\_\allowbreak{}window\_\allowbreak{}behavior \\
Random Survival Forest & active\_\allowbreak{}weeks\_\allowbreak{}first\_4 & early\_\allowbreak{}window\_\allowbreak{}behavior \\
Neural-MTLR & active\_\allowbreak{}weeks\_\allowbreak{}first\_4 & early\_\allowbreak{}window\_\allowbreak{}behavior \\
\addlinespace[4pt]
\multicolumn{3}{l}{\textit{Family~A}} \\
\addlinespace[2pt]
Linear Discrete-Time Hazard & n\_\allowbreak{}vle\_\allowbreak{}rows\_\allowbreak{}week & dynamic\_\allowbreak{}temporal\_\allowbreak{}behavioral \\
Poisson Piecewise-Exponential & n\_\allowbreak{}vle\_\allowbreak{}rows\_\allowbreak{}week & dynamic\_\allowbreak{}temporal\_\allowbreak{}behavioral \\
Neural Discrete-Time Survival & week & discrete\_time\_index \\
Gradient-Boosted Weekly Hazard & studied\_credits & discrete\_time\_index \\
\bottomrule
\end{tabular}
\setlength{\tabcolsep}{6pt}
\label{tab:explainability_summary}
\end{table}

\begin{figure}[ht]\centering
\includegraphics[width=\linewidth]{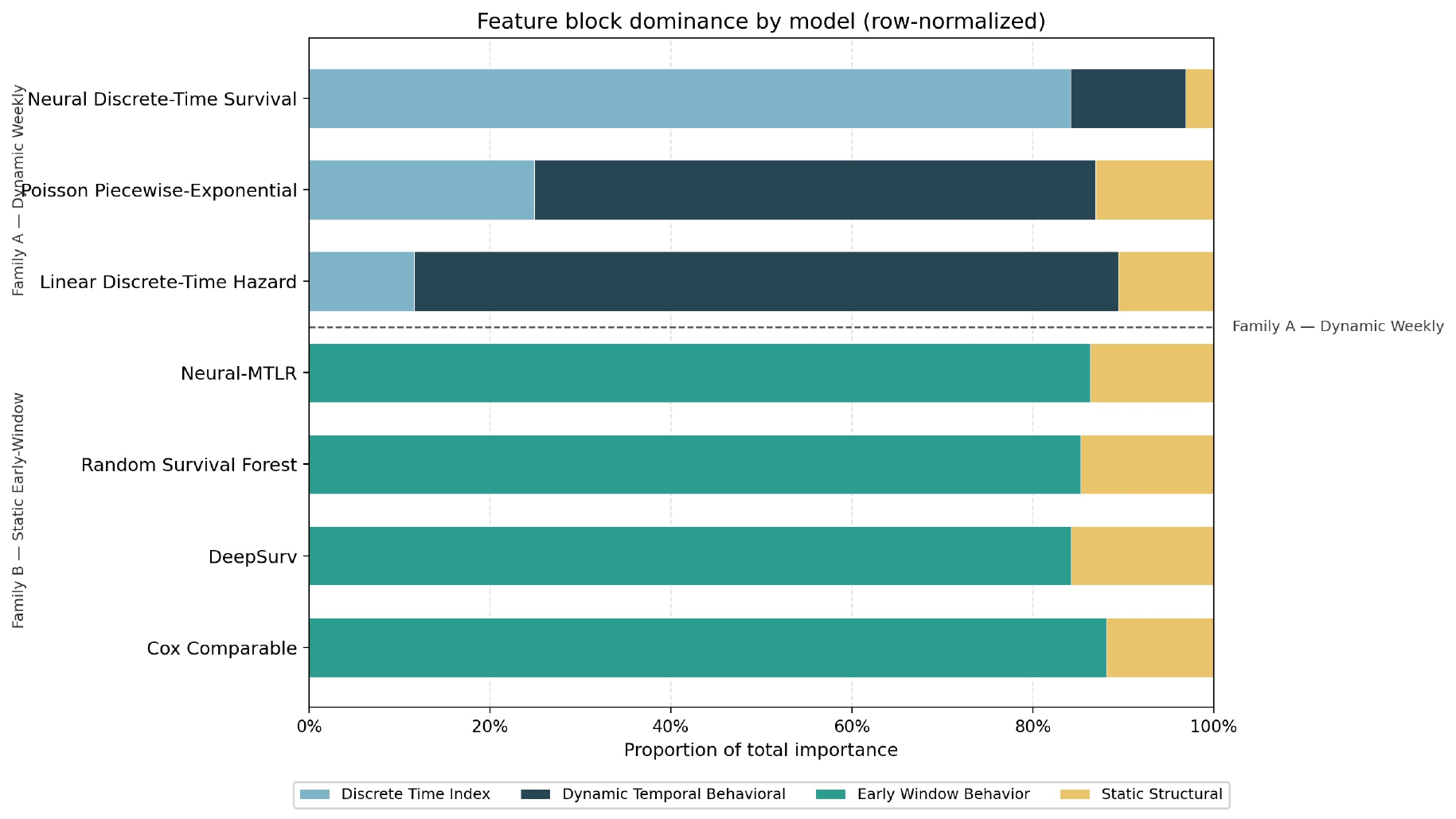}
\caption{Normalized explainability block dominance across eight models from both families included in the paper-facing export. Values are normalized within model to highlight relative block importance; the figure should be read as a within-model comparison of dominant signal blocks, not as an absolute cross-model magnitude comparison.}
\label{fig:explainability_block_dominance}
\end{figure}

Ablation and explainability converge on the same pattern: temporal or early-window signal was dominant across all eight representative models, regardless of family or architecture.

\subsection{Calibration results}

Calibration was evaluated through a quantile-bin procedure on predicted event risk at the benchmark horizons, using IPCW-adjusted empirical event rates in each bin. The reported values are the horizon-wise weighted absolute gaps defined in Equation~\ref{eq:calibration_gap}. This metric is the primary calibration criterion; slope and intercept diagnostics from a weighted reliability-bin fit are treated as strengthening auxiliaries, not as primary outputs.

Calibration results for both families are reported in Table~\ref{tab:calibration_summary}. For Family~B, predictions are generated from fixed enrollment-level representations and horizon risk is extracted directly from the predicted survival curve. For Family~A, weekly hazard predictions are accumulated into a survival curve as in Equation~\ref{eq:discrete_survival_reconstruction}, and fixed-horizon cumulative risk is extracted as in Equation~\ref{eq:discrete_risk_from_survival}; the evaluation uses study-level enrollment durations so that students who do not withdraw remain evaluable as survivors, preserving the full test set across all enrollment windows.

Within Family~B, \textit{Random Survival Forest} showed the best calibration at horizon 10 and the most consistently strong calibration profile across all three horizons (gaps: 0.012, 0.016, 0.014); most other models showed moderate calibration with gaps ranging from 0.011 to 0.044 (excluding Neural-MTLR, which was wider). \textit{XGBoost AFT} was a clear exception, with calibration gaps of 0.124, 0.178, and 0.218 at horizons 10, 20, and 30, substantially larger than any other Family~B model. \textit{DeepSurv} and \textit{Cox Comparable} showed the lowest gaps at horizons 20 and 30, respectively; \textit{Neural-MTLR} showed wider gaps across all horizons. Within Family~A, all five models showed calibration gaps between 0.089 and 0.098 at horizon 10, with \textit{Gradient-Boosted Weekly Hazard} lowest at 0.089; gaps increased with horizon, ranging from 0.123 to 0.149 at horizon 30. Family~A calibration gaps are larger in absolute magnitude than those of the best Family~B models, but they remain meaningfully below the XGBoost AFT outlier at all horizons.

\begin{table}[hbt]
\caption{Calibration gap summary for Family~A and Family~B models. Values are IPCW-adjusted mean absolute gaps between predicted risk and observed event rate across quantile bins; lower values indicate better calibration.}
\centering
\scriptsize
\setlength{\tabcolsep}{5pt}
\begin{tabular}{>{ \raggedright\arraybackslash}p{4.2cm}ccc}
\toprule
Model & Calib@10 & Calib@20 & Calib@30 \\
\midrule
\multicolumn{4}{l}{\textit{Family~B}} \\
\addlinespace[2pt]
Random Survival Forest & 0.0121 & 0.0162 & 0.0138 \\
DeepSurv & 0.0320 & 0.0145 & 0.0119 \\
DeepHit & 0.0333 & 0.0322 & 0.0270 \\
Weibull AFT & 0.0369 & 0.0219 & 0.0123 \\
Neural-MTLR & 0.0384 & 0.0503 & 0.0521 \\
Cox Comparable & 0.0386 & 0.0183 & 0.0106 \\
Gradient-Boosted Cox & 0.0437 & 0.0246 & 0.0160 \\
Royston-Parmar & 0.0438 & 0.0319 & 0.0212 \\
XGBoost AFT & 0.1242 & 0.1775 & 0.2176 \\
\addlinespace[4pt]
\multicolumn{4}{l}{\textit{Family~A}} \\
\addlinespace[2pt]
Gradient-Boosted Weekly Hazard & 0.0886 & 0.1122 & 0.1438 \\
Linear Discrete-Time Hazard & 0.0890 & 0.1123 & 0.1427 \\
Neural Discrete-Time Survival & 0.0915 & 0.1174 & 0.1490 \\
CatBoost Weekly Hazard & 0.0919 & 0.1161 & 0.1475 \\
Poisson Piecewise-Exponential & 0.0980 & 0.1023 & 0.1228 \\
\bottomrule
\end{tabular}
\setlength{\tabcolsep}{6pt}
\label{tab:calibration_summary}
\end{table}

A class-weighting sensitivity experiment (Methods) confirmed the unweighted protocol as the better-calibrated design choice. A horizon-level comparison of calibration gaps and reliability diagrams are presented in the Appendix.

\section{Discussion}

\subsection{Benchmark hierarchy and methodological scope}

Within Family~B, RSF showed consistently the best point estimate across all reported metrics. This profile was maintained under bootstrap resampling: RSF led time-dependent concordance in 90\% of resamples and maintained Brier-horizon leadership; RSF also held a 64\% IBS rank-1 share, a directional signal that should not be read as strict dominance given overlapping confidence intervals---Neural-MTLR captured 27\% of rank-1 IBS draws under resampling, indicating the IBS hierarchy is partially shared between the two top models. The within-family hierarchy is therefore largely stable rather than perfectly rigid. For practitioners deploying early-warning systems, consistent multi-metric leadership across concordance and Brier horizons provides a more reliable basis for allocating support resources than single-metric optimisation; RSF's profile across all four metrics, maintained under bootstrap resampling, makes it the more actionable choice within Family~B. Within Family~A, the tight IBS band means no single model holds a stable bootstrap lead.

Two scope conditions bound the interpretation of these results. The proportional-hazards audit classified the comparable Cox model as broad departure from proportionality (10 of 38~covariates showing possible non-proportionality at $\alpha = 0.05$), establishing that this anchor is an approximate specification rather than a fully assumption-clean Cox model; all results from this model should be read with this approximation in scope. The formal PH audit applies only to the comparable Cox anchor and should not be read as a symmetric diagnostic pass for all Family~B models. Additionally, the benchmark evaluates generalization across enrollments under shared curricular context---modules, presentations, and module-presentation combinations overlap completely across splits---not transportability to unseen curricular settings.

The result that \textit{Random Survival Forest} led on time-dependent concordance and horizon-specific Brier scores within Family~B suggests that non-parametric flexibility was sufficient to capture the dominant enrollment-level patterns in the early-window representation, even relative to the neural static early-window models tested here. The two families are best read as complementary methodological evidence about the same dropout risk phenomenon, not as a ranked comparison of weekly versus early-window modelling strategies.

\subsection{Feature signal attribution}

The ablation experiment localizes the source of predictive performance. The main gains were not sustained by relatively stable background attributes; the most severe losses emerged when models were deprived of variables representing activity, progression, and engagement across the course. In this dataset, dropout risk does not appear primarily as a background trait but as a pattern that becomes visible through the temporal trajectory of academic and behavioral participation \citep{Mubarak2020InteractionLogs,Vaarma2024FinnishHE}. That engagement features constructed as primary indicators do, in fact, dominate is a necessary validity check rather than a surprising structural claim: it confirms the pipeline captures the intended temporal signal and is not driven by noise or by an unintended proxy. Read this way, the result is a control that licenses the subsequent cross-model comparison, not a deflation of it. Such a reading aligns with a Learning Analytics perspective in which prediction's analytical value lies not only in classification but in surfacing temporally meaningful signals of persistence, disengagement, and possible withdrawal \citep{Ifenthaler2020StudySuccess,Alalawi2024SPPA}.

\begin{sloppypar}
The top individual drivers for the Family~B models are behavioral summaries from the early observation window: \textit{clicks\_first\_4\_weeks} for Cox Comparable and \textit{active\_weeks\_first\_4} for DeepSurv, RSF, and Neural-MTLR. These variables summarize observable early engagement that covaries with dropout risk, but do not by themselves support causal claims about why dropout occurs \citep{Arnold2020PrognosisIntervention,Hooshyar2024SHAPLIME}.
\end{sloppypar}

These ablation and explainability findings are paradigm-agnostic: all eight models assigned dominance to a temporal block regardless of family or architecture, with both analyses converging on the same directional conclusion across linear, tree-based, and neural paradigms \citep{Fisher2019AllModelsWrong}.

\subsection{Calibration coherence and risk structure}

Risk models can produce strong discrimination while generating unreliable probability estimates \citep{VanCalster2019CalibrationAchilles,Kvamme2019TimeToEvent}. In the present benchmark, calibration reinforced rather than challenged the main benchmark hierarchy: within Family~B, \textit{RSF} showed the best calibration at horizon 10 and the most consistently strong calibration profile across all three horizons (gaps: 0.0121, 0.0162, 0.0138), confirming that its discrimination advantage is not accompanied by calibration coherence loss. The \textit{XGBoost AFT} outlier also aligns with its benchmark outlier status on IBS and TD concordance, suggesting that its overall weaker result reflects a consistent estimation failure rather than an isolated metric artefact. From a risk-structure perspective, the miscalibration of XGBoost AFT's time-invariant AFT parameterisation while more flexible hazard formulations (RSF, DeepSurv, Cox) remain better-calibrated suggests that dropout hazard ratios in this dataset are temporally heterogeneous; the miscalibration is therefore a structural diagnostic about the risk process, indicating that dropout risk does not unfold with time-constant covariate acceleration, and not merely a basis for model selection. Within Family~A, all five models produced calibration gaps below 0.10 at horizon 10, substantially smaller than the XGBoost AFT outlier, confirming that the Family~A weekly hazard predictions, when read as horizon-specific cumulative risk estimates, are numerically coherent with observed dropout rates. Family~A gaps are larger in absolute magnitude than those of the best Family~B models, though substantially below the XGBoost AFT outlier.

\subsection{Synthesis}

Benchmark, ablation, explainability, calibration, and bootstrap evidence converged on a consistent interpretation. The main comparative conclusions do not rest on a single metric or on a single reading of performance. Rather, they emerge from the alignment among global probabilistic error, time-dependent discrimination, horizon-specific error, and calibration coherence (evaluated within each family separately, and supported by enrollment-level bootstrap resampling as a ranking-stability diagnostic).

The answers to the three research questions reinforced rather than contradicted each other, with one important qualification for each. For RQ1, the benchmark yielded a largely stable within-family hierarchy: within Family~B, \textit{Random Survival Forest} led on time-dependent concordance (0.674; 90\% rank-1 bootstrap share, largely stable) and on all three Brier horizons; within Family~A, \textit{Poisson Piecewise-Exponential} led narrowly on IBS (0.1399) within a tight five-model cluster; IBS bootstrap leadership was directional (RSF rank-1 share 64\%; Neural-MTLR 27\%), though intervals still overlap and the ranking should not be treated as strictly dominant. The ordering within each family is consistent but not strictly dominant: both families show modest absolute margins. Family~A and Family~B cannot be pooled into a single global ranking without erasing the representational difference between weekly discrete-time hazard and fixed early-window continuous-time survival predictions.

For RQ2, the systematic ablation on eight representative models (four per family) showed that temporal-behavioral signal removal caused substantially larger performance loss than static removal in all Family~B models (IBS ratio range: 1.20--2.46) and on concordance across all eight models; this finding holds regardless of modelling paradigm.

For RQ3, the paper-facing explainability export (covering eight models across both families) found that the dominant feature block in all eight models is temporal (\textit{dynamic\_temporal\_behavioral}, \textit{discrete\_time\_index}, or \textit{early\_window\_behavior}), a result that now converges with rather than complements the ablation evidence. Calibration within Family~B showed that \textit{RSF} had the best calibration profile (gaps: 0.012--0.016), most other models showed moderate gaps (0.011--0.044, excluding Neural-MTLR, which was wider), with \textit{XGBoost AFT} as a consistent extreme outlier across all metrics; Family~A models showed calibration gaps between 0.089 and 0.098 at horizon 10, confirming they generate numerically coherent probability estimates rather than degenerate risk scores. Together, explainability and calibration converge on a consistent answer to RQ3: dropout risk in this dataset is structured as a temporally evolving process driven primarily by behavioral engagement signals, with a risk function amenable to well-calibrated survival modeling when the chosen formulation is sufficiently flexible to accommodate time-varying hazard dynamics.

Methodologically, these results suggest that comparisons between temporal dropout risk models in Learning Analytics may benefit from unified, multi-dimensional protocols, as opposed to isolated metrics or cross-study synthesis under heterogeneous settings \citep{Farhood2024ComparativeAI}. This also places the paper in direct dialogue with recent survival-evaluation arguments that caution against single-metric reading and instead recommend evaluation layers that remain explicit about what each metric validates and what it does not \citep{Lillelund2025StopChasingCIndex}. Substantively, dropout risk in this context appears as a process structured by temporal signal (early participation, activity intensity, weekly progression, and recent disengagement), and this conclusion is supported consistently by both ablation and explainability analyses, with both converging on temporal dominance across all eight representative models and both families \citep{Marcolino2025MoodleLogs}.

The leading position of Random Survival Forest in Family~B is consistent with findings from neutral survival benchmarks on non-educational datasets. Kvamme~et~al.~\citep{Kvamme2019TimeToEvent} evaluated RSF, Cox, DeepSurv, and DeepHit across several standard clinical survival datasets and found RSF to be persistently competitive, though model leadership varied by dataset and scoring rule. Burk~et~al.~\citep{Herrmann2021SurvivalBenchmark} conducted a large-scale neutral comparison of survival methods on low-dimensional data and likewise found tree-based ensembles to be competitive with the Cox baseline, particularly in datasets with moderate to strong signal-to-noise ratios. The consistency between the present OULAD result and these external references strengthens the interpretation that RSF's discrimination advantage reflects algorithmic characteristics rather than dataset-specific artifact. That said, model rankings in those neutral benchmarks varied across datasets and metrics, indicating that replication of the present benchmark on other educational datasets remains a necessary step before generalizing the Family~B hierarchy beyond the shared-context OULAD setting.

All findings are bounded by shared curricular context (all modules and presentations appear in both splits), and bootstrap intervals reflect enrollment-sampling variance only, not a formal hypothesis test or retraining-based generalization claim.

\paragraph*{Takeaway.}
In one line: under a unified protocol, 14~models across two representations agree that dropout is a temporal-behavioral process, and within the early-window family the tree-based Random Survival Forest leads on a multi-metric basis (TD concordance 0.674, IBS 0.1118) with a 90\% bootstrap rank-1 share on discrimination. What the benchmark does \emph{not} establish: transportability to unseen modules/presentations, or formal hypothesis tests of the rankings (the no-refit bootstrap bounds sampling variance only). For a practitioner: prefer RSF within the early-window family on the strength of consistent multi-metric leadership, not a single score.

\subsection{Practical implications}

Three conclusions from the benchmark carry a direct operational reading for institutions that collect LMS engagement data.

First, early-window prediction is viable at week four. RSF trained on the first four weeks of engagement consistently outperformed alternatives on discrimination and calibration. An institution with access to click counts and weekly activity logs in its VLE can produce calibrated dropout-risk scores at week~4, before the critical withdrawal window in most modules. Because RSF's probabilities are well-calibrated (Family~B gaps: 0.012--0.016), a threshold such as $P(\text{dropout}) > 0.3$ corresponds to a meaningful risk level, not an arbitrary score rank. A practical three-step protocol follows from these properties: score all active enrolments at the end of week~4 using the first-four-week behavioral summary; flag those with $\hat{P}(\text{dropout}) > 0.30$ for proactive tutor contact; and, among flagged students, prioritise those with simultaneously low VLE activity and few active weeks---the dominant SHAP contributors across all Family~B models. The lower IBS of RSF translates directly into this threshold being closer to the empirical withdrawal rate, making $0.30$ operationally meaningful rather than a percentile cut.

Second, the dominant predictive signal is early engagement, not background attributes. Ablation and explainability results agree: what predicts dropout is observable in-course behavior---clicks in the first four weeks and number of active weeks---rather than demographic background or prior qualifications. Institutions do not need complex data pipelines; the most predictive variables are the simplest to collect from any LMS.

Third, weekly monitoring adds value under a different deployment contract. Family~A models can, in principle, update risk scores throughout the course rather than only at week~4. The trade-off is weaker enrollment-level calibration relative to Family~B. A practical hybrid would use an early-window RSF score at week~4 to trigger initial outreach, then use weekly engagement trends as a qualitative signal for ongoing case management, without treating weekly model scores as calibrated probabilities.

One constraint applies across all three readings: the benchmark was conducted under shared curricular context, and deployment on unseen modules or different institutional settings requires local revalidation before the reported metrics can be treated as expected performance bounds \citep{Sonderlund2018LAInterventions}.

\section{Conclusion}

This study developed and applied a unified survival-oriented benchmark for temporal dropout risk modelling in Learning Analytics, comparing 14 tuned models organized into two methodologically distinct arms: a Family~A of five weekly discrete-time hazard models and a Family~B of nine early-window continuous-time survival models. Its purpose was not only to compare models, but to make those comparisons methodologically clearer and educationally interpretable under a shared evaluation protocol.

Within Family~B, \textit{Random Survival Forest} led on time-dependent concordance (0.674) and on all three reported Brier horizons, with bootstrap evidence supporting a largely stable rank-1 position on discrimination across 90\% of enrollment-level resamples. \textit{RSF} showed the best calibration profile within Family~B; \textit{XGBoost AFT} was a consistent extreme outlier across all metrics. Within Family~A, the five models clustered within a 0.0011 IBS band, with \textit{Poisson Piecewise-Exponential} narrowly leading. The two families use different risk formulations and evaluation footprints, so that a single cross-family ranking would conflate genuine model differences with representational artifacts rather than measure real differences. A systematic ablation on eight representative models (four per family) showed that temporal-behavioral signal removal caused larger performance loss than static removal in all Family~B models (IBS ratio range: 1.20--2.46) and on concordance across all eight models. The paper-facing explainability export (covering eight models across both families) identified a temporal or early-window block as dominant in all eight models, a finding that converges with rather than merely complements the ablation evidence.

Two contributions stand out. Methodologically, the study shows the value of comparing temporal dropout risk models under a unified, multi-dimensional protocol organized by family group, rather than through isolated metrics or indirect synthesis across heterogeneous studies \citep{Dwivedi2022BenchmarkingGNNs,Collins2024Evaluation}. Substantively, the ablation and explainability results converge on a paradigm-agnostic conclusion: temporal-behavioral signal (whether captured as weekly activity in Family~A or early-window engagement summaries in Family~B) dominates both the aggregate-removal impact and the block-level attribution across all eight representative models and both families. The benchmark structure itself is the transferable contribution: the OULAD provided a standardised substrate for its demonstration, not an end in itself.

\subsection*{Limitations}\label{sec:limitations}

Three scope qualifications bound the interpretation of all reported findings: (i)~external generalizability is constrained by the single-dataset design; (ii)~bootstrap confidence intervals are no-refit lower bounds on ranking uncertainty, reflecting enrollment-sampling variance only and not the additional estimation variance from model refitting or hyperparameter re-selection; and (iii)~post-hoc recalibration was not applied to the held-out predictions.

The benchmark is based on a single dataset and should not be treated as automatically generalizable to other institutional settings. The split evaluated generalization under shared curricular context (all 7~modules and 4~presentations appear in both partitions; Table~\ref{tab:appendix_split_context_audit}) rather than module- or presentation-level transportability, and bootstrap intervals reflect enrollment-sampling variance only. The early-window choice ($w=4$) was validated against a sensitivity grid ($w \in \{2,4,6,8,10\}$ weeks; Table~\ref{tab:appendix_window_sensitivity}). Eight directions remain as future work: decision-analytic evaluation (decision curves, net benefit); fairness and subgroup-performance auditing; comparison to non-survival LA/EDM baselines (e.g., logistic regression, decision trees); cross-institutional replication; recalibration experiments for Family~A models (reliability diagrams in Figures~\ref{fig:appendix_reliability_comparable_a}--\ref{fig:appendix_reliability_dynamic}); cross-module transportability studies \citep{Sonderlund2018LAInterventions,Alalawi2024SPPA}; extension to time-varying covariate designs via landmarking; and ensemble combinations across models.

\FloatBarrier

\begin{appendices}

\section{Appendix}

\subsection{Evaluation protocol audit}

Table~\ref{tab:appendix_protocol_audit} summarizes the main operational conventions used in the benchmark. Its purpose is not to introduce a second evaluation framework, but to make explicit the conventions already used in the empirical pipeline: the enrollment-level unit of analysis, the event definition, the censoring treatment for IBS and horizon-specific Brier score, the concordance convention reported in the benchmark tables, the distinction between dynamic and static early-window prediction settings, the enrollment-level identity leakage result, the contextual scope of the split, the operational definition of horizon-specific calibration, the inferential role of bootstrap, and the proportional-hazards scope boundary.

\begin{table}[!t]
\caption{Evaluation protocol audit for the unified benchmark.}
\centering
\scriptsize
\setlength{\tabcolsep}{2pt}
\renewcommand{\arraystretch}{1.12}
\begin{tabularx}{\linewidth}{@{}>{ \raggedright\arraybackslash}p{2.8cm}
                              >{\raggedright\arraybackslash}p{3.6cm}
                              >{\raggedright\arraybackslash}X@{}}
\toprule
Component & Status/value & Details \\
\midrule
Unit of analysis & defined: enrollment & All final benchmark comparisons are reported at the enrollment level. \\
Event definition & \makecell[l]{defined: withdrawn\\with valid date\\(unregistration)} & Observed withdrawal event with a valid time stamp (unregistration with a recorded date). \\
Official horizons & defined: 10, 20, 30 & Shared benchmark horizons used for Brier, IBS, and calibration reporting. \\
Primary discrimination metric & \makecell[l]{defined: TD\\concordance} & The canonical discrimination metric is time-dependent concordance rather than a static concordance proxy. \\
Censoring treatment & \makecell[l]{defined: IPCW\\with KM} & Brier score and IBS use inverse-probability-of-censoring weighting with the Kaplan--Meier censoring estimator. \\
Dynamic versus comparable rule & \makecell[l]{defined: dynamic weekly\\vs early-window comparable} & Weekly discrete-time arms update predictions over person-period rows, whereas comparable continuous-time arms use early-window enrollment representations; cross-family comparison occurs only after shared enrollment-level horizon reporting. \\
Leakage prevention rule & \makecell[l]{defined: no enrollment\\identity leakage} & The split is enforced at the enrollment level with no identity leakage between train and test. \\
Split scope boundary & \makecell[l]{defined: shared\\curricular context} & The benchmark generalizes across enrollments under shared curricular context rather than under context-disjoint transportability conditions. \\
Primary calibration metric & \makecell[l]{defined: weighted absolute\\gap by horizon} & The main calibration criterion is the weighted absolute calibration gap at each benchmark horizon. \\
Calibration strengthening & \makecell[l]{defined: intercept\\and slope by horizon} & Intercept and slope are retained as strengthening diagnostics rather than as the primary benchmark ranking criterion. \\
Bootstrap inferential role & \makecell[l]{defined: ranking support,\\not formal testing} & Bootstrap is used to assess how stable the exported ranking appears, not to claim a formal hypothesis-testing result. \\
PH scope boundary & \makecell[l]{defined: formal PH audit\\for Cox anchor only} & Formal classical proportional-hazards auditing is available for the comparable Cox anchor, but not in an equivalent form for DeepSurv. \\
\bottomrule
\end{tabularx}
\setlength{\tabcolsep}{6pt}
\renewcommand{\arraystretch}{1.0}
\label{tab:appendix_protocol_audit}
\end{table}

\subsection{Calibration evidence by horizon}
\label{sec:calib_evidence}

To complement Table~\ref{tab:calibration_summary}, the appendix retains the horizon-wise calibration evidence for Family~B. In the canonical final export, no single model held the lowest calibration gap at all three horizons: \textit{Random Survival Forest} led at horizon 10 (0.0121), \textit{DeepSurv} at horizon 20 (0.0145), and \textit{Cox Comparable} at horizon 30 (0.0106). \textit{XGBoost AFT} remained the only notable exception, with gaps of 0.1242, 0.1775, and 0.2176 at horizons 10, 20, and 30, consistent with its outlier status across benchmark performance and discrimination metrics. Family~A calibration gaps at horizon 10 ranged from 0.089 (\textit{Gradient-Boosted Weekly Hazard}) to 0.098 (\textit{Poisson Piecewise-Exponential}); these are reported in Table~\ref{tab:calibration_summary} and confirm that all five Family~A models generate numerically coherent horizon-specific risk estimates.

Slope and intercept estimates are retained as strengthening diagnostics only, not as a second parallel ranking criterion. In the consolidated interpretation, RSF, which led on time-dependent concordance and Brier horizons, also showed the best calibration profile among all Family~B models (gaps: 0.0121, 0.0162, 0.0138), supporting the main benchmark hierarchy without contradiction.

Accordingly, the appendix uses the tabulated horizon-wise calibration evidence as the primary robustness material for calibration, without a separate figure.

\subsection{Preprocessing and tuning audit}

Table~\ref{tab:appendix_preproc_tuning_audit} summarizes the preprocessing and tuning conventions for the four Family~B models retained in the focused tuning audit. All four used the same upstream preprocessing discipline: median imputation for numeric variables, constant-missing category for categorical variables, one-hot encoding with unknown handling, and standard scaling fitted exclusively on training rows. The remaining ten tuned families (three in Family~A and six additional Family~B models) applied the same preprocessing contract; their individual candidate records are documented in the pipeline scripts and available in the open repository. A machine-readable summary of all search grids is provided as Online Resource~1 (\texttt{ESM\_1.pdf}), auto-generated from pipeline run-time metadata by \path{dropout_bench_v3_G_13_supplementary_hyperparameter_grids.py}.

\begin{table}[!t]
\caption{Preprocessing and tuning audit for the four Family~B models retained in the focused comparative export. All families used median numeric imputation, constant-missing categorical imputation, one-hot encoding, and standard scaling fit on training rows only. Tuning was deliberately controlled, not exhaustive.}
\centering
\tiny
\setlength{\tabcolsep}{2pt}
\renewcommand{\arraystretch}{1.08}
\begin{tabularx}{\linewidth}{@{}>{\raggedright\arraybackslash}p{2.0cm}
                                >{ \raggedright\arraybackslash}p{1.6cm}
                                >{\raggedright\arraybackslash}X
                                >{\raggedright\arraybackslash}X@{}}
\toprule
Model & Input level & Validation strategy & Tuning and complexity control \\
\midrule
Cox Comparable &
\makecell[tl]{enrollment\\early window} &
Enrollment-level split with event stratification when possible (20\%). &
24 candidates. Selection by highest validation C-index. Restricted regularization grid over penalizer and \(l_1\)-ratio. No early stopping. \\

DeepSurv &
\makecell[tl]{enrollment\\early window} &
Internal validation fraction on training rows (20\%). &
24 candidates. Selection by lowest validation loss. Early stopping (patience 10). Complexity controls: architecture grid, dropout, weight decay, best-epoch refit. \\

Random Survival Forest &
\makecell[tl]{enrollment\\early window} &
Enrollment-level split with event stratification when possible (20\%). &
24 candidates. Selection by lowest validation IBS. No early stopping. Complexity controls: forest size, minimum leaf size, maximum depth, feature subsampling fraction. \\

Neural-MTLR &
\makecell[tl]{enrollment\\early window} &
Enrollment-level split with deterministic survival-time discretization (10\%). &
24 candidates. Selection by lowest validation IBS. Early stopping (patience 8). Complexity controls: discretization grid, architecture, dropout, weight decay. \\
\bottomrule
\end{tabularx}
\setlength{\tabcolsep}{6pt}
\renewcommand{\arraystretch}{1.0}
\label{tab:appendix_preproc_tuning_audit}
\end{table}

\textit{XGBoost AFT}, the consistent outlier across IBS, TD concordance, and calibration, applied the same upstream preprocessing discipline and was tuned over 24 candidates varying AFT loss distribution (logistic, normal), distribution scale (1.0, 1.5, 2.0), learning rate (0.03--0.08), and boosting rounds (200--400); tree depth was held fixed at \(\texttt{max\_depth}=2\) throughout the grid. A plausible hypothesis for its systematic underperformance is the adverse interaction between the AFT parametrization (which assumes time-invariant multiplicative covariate effects on the acceleration factor) and the broad proportional-hazards departure observed in this dataset (26.3\% of covariates showed evidence of non-pro\-por\-tion\-al\-i\-ty in the comparable Cox audit, classified as broad departure from proportionality). When relative hazard ratios exhibit strong temporal heterogeneity, even a gradient-boosted ensemble cannot fully compensate for the misspecification introduced by a fixed AFT time-to-event parametrization, and the persistence of this gap under the expanded 24-candidate search reinforces a structural rather than a tuning-related explanation. The failure mode has not been diagnosed beyond this structural hypothesis; formal diagnosis would require analysis of predicted survival curve shapes, AFT residuals, and risk score distributions across event-time strata, which remain outside the current benchmark scope.

\subsection{Bootstrap uncertainty for the benchmark hierarchy}

Because the leading performance differences were modest in absolute magnitude, uncertainty was quantified on the held-out test set through enrollment-level bootstrap resampling (200~resamples, no model refit within iterations, fixed frozen survival predictions). The bootstrap covers the four Family~B models retained in the focused comparative audit. Family~A families are excluded because their weekly discrete-time hazard predictions are not directly comparable under the same fixed-horizon survival evaluation footprint. Table~\ref{tab:appendix_bootstrap_uncertainty} reports the point estimate and 95\% bootstrap interval for IBS, time-dependent concordance, and the three Brier horizons.

\begin{table}[!t]
\caption{Bootstrap uncertainty for the four Family~B models (200 enrollment-level resamples, no refit, frozen survival predictions). Lower IBS and Brier scores indicate better probabilistic accuracy (lower mean squared survival error); higher time-dependent concordance indicates better discrimination.}
\centering
\scriptsize
\setlength{\tabcolsep}{3pt}
\renewcommand{\arraystretch}{1.1}
\begin{tabular}{>{\raggedright\arraybackslash}p{2.6cm} >{\centering\arraybackslash}p{2.0cm} >{\centering\arraybackslash}p{2.0cm} >{\centering\arraybackslash}p{1.5cm} >{\centering\arraybackslash}p{1.5cm} >{\centering\arraybackslash}p{1.5cm}}
\toprule
Model & IBS [95\% CI] & TD concordance [95\% CI] & Brier@10 [95\% CI] & Brier@20 [95\% CI] & Brier@30 [95\% CI] \\
\midrule
RSF & 0.1118 [0.109, 0.115] & 0.6735 [0.663, 0.684] & 0.0947 [0.091, 0.099] & 0.1329 [0.129, 0.137] & 0.1574 [0.153, 0.161] \\
DeepSurv & 0.1145 [0.111, 0.118] & 0.6619 [0.651, 0.673] & 0.0978 [0.094, 0.102] & 0.1343 [0.131, 0.138] & 0.1589 [0.155, 0.163] \\
Cox Comparable & 0.1176 [0.114, 0.121] & 0.6512 [0.640, 0.663] & 0.1012 [0.097, 0.105] & 0.1378 [0.133, 0.142] & 0.1623 [0.158, 0.166] \\
Neural-MTLR & 0.1131 [0.109, 0.117] & 0.6569 [0.647, 0.668] & 0.0954 [0.091, 0.100] & 0.1340 [0.129, 0.139] & 0.1588 [0.154, 0.163] \\
\bottomrule
\end{tabular}
\setlength{\tabcolsep}{6pt}
\renewcommand{\arraystretch}{1.0}
\label{tab:appendix_bootstrap_uncertainty}
\end{table}

The bootstrap results confirm that margins are modest in absolute terms: IBS confidence intervals for the top four models overlap substantially, so the IBS hierarchy should be treated as a directional signal rather than a stable dominance claim. On time-dependent concordance, RSF maintained rank~1 in 90\% of resamples, a largely stable leadership pattern. These bootstrap artifacts do not constitute a formal hypothesis test and should not be narrated as unqualified strict superiority.

\subsection{Bootstrap inferential scope}
\label{subsec:bootstrap_scope}

Table~\ref{tab:appendix_bootstrap_scope} clarifies what the bootstrap artifacts do and do not support in the current paper. On time-dependent concordance, \textit{Random Survival Forest} held rank~1 in 90\% of resamples, a largely stable leadership pattern, with \textit{DeepSurv} as the main runner-up (10\% rank-1 share). On IBS, RSF led with a rank-1 share of 64\%; the IBS hierarchy is directional but the intervals overlap, and the ranking should not be treated as strictly dominant (RSF rank-1 share: 64\%; Neural-MTLR: 27\%).

The inferential reading should remain disciplined. These bootstrap artifacts support cautious directional language within the Family~B audit subset, but they do not constitute a formal null-hypothesis test, do not establish a universally fixed ranking across samples, and should not be narrated as unqualified strict superiority. Because the resampling procedure uses frozen survival predictions without refitting the model within each bootstrap iteration, the reported intervals capture enrollment-sampling variance only, not the estimation variance of the fitted model parameters. The intervals should therefore be interpreted as conservative lower-bound estimates of total uncertainty, not as full inferential confidence intervals for the generalization performance of the models.

\begin{table}[!t]
\caption{Bootstrap inferential scope for the main benchmark leadership claims (Family~B, 200 enrollment-level resamples, no refit).}
\centering
\scriptsize
\setlength{\tabcolsep}{4pt}
\renewcommand{\arraystretch}{1.08}
\begin{tabularx}{\linewidth}{@{}>{\raggedright\arraybackslash}p{2.3cm} >{\raggedright\arraybackslash}p{1.5cm} >{\raggedright\arraybackslash}p{2.4cm} >{\raggedright\arraybackslash}p{2.0cm} X@{}}
\toprule
Metric & Direction & Leading model & Claim status & Conservative takeaway \\
\midrule
Integrated Brier Score & lower is better & RSF & directional signal only & IBS intervals overlap substantially; RSF leads narrowly but the ranking is not stable enough for a strong dominance claim. \\
Time-dependent concordance & higher is better & RSF & largely stable bootstrap leadership & RSF held rank~1 in 90\% of resamples; largely stable but not absolute leadership. \\
\bottomrule
\end{tabularx}
\setlength{\tabcolsep}{6pt}
\renewcommand{\arraystretch}{1.0}
\label{tab:appendix_bootstrap_scope}
\end{table}

\subsection{Split and contextual overlap audit}

Because the benchmark uses an enrollment-level train--test split, we audited not only enrollment identity leakage but also the degree of shared curricular context across the partition. This subsection clarifies the scope of the benchmark evaluation: it should be read as generalization across enrollments under shared curricular context, not as a stricter transportability setting across unseen module or presentation contexts.

Table~\ref{tab:appendix_split_context_audit} summarizes the compact split/context audit exported from the notebook. The benchmark used 32,593~enrollments in total, with 22,815 in train and 9,778 in test. Event rates were closely matched across the two partitions, and no enrollment identity leakage was detected. At the same time, curricular context was fully shared across train and test: all 7~modules, all 4~presentations, and all 22 module-presentation combinations appeared in both splits. The benchmark should therefore be interpreted as enrollment-level evaluation under shared curricular context, not as a context-disjoint transportability benchmark.

\begin{table}[!t]
\caption{Split and contextual overlap audit for the main benchmark.}
\centering
\scriptsize
\setlength{\tabcolsep}{4pt}
\renewcommand{\arraystretch}{1.1}
\begin{tabular}{>{\raggedright\arraybackslash}p{1.8cm} >{\raggedright\arraybackslash}p{2.9cm} >{\centering\arraybackslash}p{1.2cm} >{\centering\arraybackslash}p{1.2cm} >{\centering\arraybackslash}p{1.2cm} >{\centering\arraybackslash}p{1.1cm} >{\centering\arraybackslash}p{1.1cm}}
\toprule
Split unit & Stratification & Total & Train & Test & Train event rate & Test event rate \\
\midrule
enrollment & event status + coarse event-time bucket & 32,593 & 22,815 & 9,778 & 0.2266 & 0.2266 \\
\bottomrule
\end{tabular}

\vspace{0.4em}

\begin{tabular}{>{\raggedright\arraybackslash}p{2.5cm} >{\centering\arraybackslash}p{1.6cm} >{\centering\arraybackslash}p{1.8cm} >{\centering\arraybackslash}p{2.2cm}}
\toprule
Identity leakage & Shared modules & Shared presentations & Shared module-presentations \\
\midrule
no & 7/7 & 4/4 & 22/22 \\
\bottomrule
\end{tabular}
\setlength{\tabcolsep}{6pt}
\renewcommand{\arraystretch}{1.0}
\label{tab:appendix_split_context_audit}
\end{table}

\subsection{Discrete-time diagnostic bridge}

The purpose of this subsection is to clarify why the dynamic discrete-time branch underperformed in the final enrollment-level benchmark without collapsing that result into a simplistic claim of absent weekly signal. The frozen diagnostic export shows that both discrete-time anchor models retained non-trivial row-level predictive structure, but that this local signal did not translate as effectively as the early-window comparable representation when the evaluation target shifted to enrollment-level survival ranking, error, and calibration.

Table~\ref{tab:appendix_discrete_time_summary} summarizes the key diagnostic contrast. Both dynamic models were explicitly tuned, used approximately 23.8 weekly rows per enrollment on the test set, and operated under low row event rates near 0.0095. The neural weekly model improved row-level ROC-AUC and log loss relative to the linear weekly model, yet the final time-dependent concordance ranking remained weaker. This pattern is consistent with a harder aggregation problem in which sparse low-base-rate weekly hazards must be converted into stable enrollment-level survival outputs.

\begin{table}[!t]
\caption{Discrete-time diagnostic summary for Family~A models.}
\centering
\scriptsize
\setlength{\tabcolsep}{3pt}
\renewcommand{\arraystretch}{1.08}
\begin{tabularx}{\linewidth}{@{}>{ \raggedright\arraybackslash}p{2.8cm} >{ \raggedright\arraybackslash}p{1.8cm} >{\centering\arraybackslash}p{1.0cm} >{\centering\arraybackslash}p{1.2cm} >{\centering\arraybackslash}p{1.2cm} >{\centering\arraybackslash}p{1.0cm} >{\centering\arraybackslash}p{1.0cm}@{}}
\toprule
Model & Representation & \makecell[c]{Candi-\\dates} & \makecell[c]{Rows per\\enrollment} & \makecell[c]{Row\\event\\rate} & \makecell[c]{IBS\\rank} & \makecell[c]{TD\\rank} \\
\midrule
Neural Discrete-Time & \makecell[l]{dynamic weekly\\person-period} & 24 & 23.77 & 0.0095 & 4 & 1 \\
Linear Discrete-Time & \makecell[l]{dynamic weekly\\person-period} & 24 & 23.77 & 0.0095 & 1 & 2 \\
\bottomrule
\end{tabularx}
\setlength{\tabcolsep}{6pt}
\renewcommand{\arraystretch}{1.0}
\label{tab:appendix_discrete_time_summary}
\end{table}

Table~\ref{tab:appendix_discrete_time_hypotheses} then records a conservative interpretation of the main diagnostic hypotheses considered in the final evidence audit. Weak tuning alone is not supported as the primary explanation, because both discrete-time anchor models were explicitly tuned under bounded searches. Absent weekly signal is also not supported, because row-level discrimination and loss remained informative. The best-supported residual interpretation is therefore representational inefficiency under the present benchmark contract: Family~A captures signal, but less efficiently than the early-window comparable design when the benchmark outcome is enrollment-level survival prediction.

\begin{table}[!t]
\caption{Hypothesis audit for the discrete-time underperformance reading.}
\centering
\scriptsize
\setlength{\tabcolsep}{3pt}
\renewcommand{\arraystretch}{1.08}
\begin{tabularx}{\linewidth}{@{}>{\raggedright\arraybackslash}p{2.6cm} >{\raggedright\arraybackslash}p{2.2cm} X@{}}
\toprule
Hypothesis & Status & Conservative takeaway \\
\midrule
Weak tuning as primary driver & tested, not supported as primary explanation & Family~A models were explicitly tuned; weak tuning alone is not the best-supported explanation for the final benchmark gap. \\
Absent weekly signal & tested, not supported & Weekly rows contain predictive signal; the underperformance emerges downstream when those weekly hazards must support enrollment-level survival ranking and calibration. \\
Sparse low-base-rate aggregation burden & tested, partially supported & The dynamic representation faces a sparse low-base-rate aggregation problem that is consistent with harder survival reconstruction and calibration, but this remains evidence-consistent rather than causal proof. \\
Neural weekly instability or overdispersion & tested, partially supported & Family~A neural models appear more expressive locally, but that extra flexibility does not translate into a stronger final ranking profile under the benchmark contract. \\
Representational inefficiency under the current contract & residual interpretation best supported & Family~A captures signal, but less efficiently than the early-window comparable design under the present benchmark contract. \\
\bottomrule
\end{tabularx}
\setlength{\tabcolsep}{6pt}
\renewcommand{\arraystretch}{1.0}
\label{tab:appendix_discrete_time_hypotheses}
\end{table}

\subsection{Proportional-hazards audit for the comparable Cox model}

Because the static early-window family is methodologically anchored by a Cox-type model, the comparable Cox benchmark was subjected to a formal proportional-hazards audit. The purpose of this appendix subsection is not to reopen model selection, but to clarify the diagnostic status of the classical Cox assumption within the static early-window branch.

Table~\ref{tab:appendix_cox_ph_summary} summarizes the global result of the audit. Of the 38 tested covariates, 10 showed evidence of possible non-proportionality at $\alpha = 0.05$, corresponding to 26.3\% of the tested covariates. The notebook assigns a global classification of broad departure from proportionality: the comparable Cox benchmark shows broad evidence of proportional-hazards departure and should be interpreted as an approximate comparable benchmark rather than a fully assumption-clean Cox specification.

\begin{table}[!t]
\caption{Global proportional-hazards audit summary for the comparable Cox model.}
\centering
\scriptsize
\setlength{\tabcolsep}{4pt}
\renewcommand{\arraystretch}{1.1}
\begin{tabularx}{\linewidth}{@{}>{ \raggedright\arraybackslash}p{3.0cm}
                              >{\centering\arraybackslash}p{1.6cm}
                              >{\centering\arraybackslash}p{1.6cm}
                              >{\raggedright\arraybackslash}X@{}}
\toprule
Model & Covariates tested & Flagged & Global interpretation \\
\midrule
Cox Comparable & 38 & 10 (26.3\%) & Broad departure: approximate comparable benchmark, not assumption-clean Cox specification \\
\bottomrule
\end{tabularx}
\setlength{\tabcolsep}{6pt}
\renewcommand{\arraystretch}{1.0}
\label{tab:appendix_cox_ph_summary}
\end{table}

The covariates with the strongest evidence of possible non-proportionality in the current evidence freeze were the following:

\begin{enumerate}[label=(\roman*), leftmargin=*]
\item \texttt{region} (level: Scotland)
\item \texttt{studied\_credits}
\item \texttt{region} (level: Wales)
\item \texttt{age\_band} (level: 0--35)
\item \texttt{region} (level: Ireland)
\end{enumerate}

The enumerated covariates above were identified as the strongest signals of non-pro\-por\-tion\-al\-i\-ty, confirming the broad departure from proportionality classification.

This audit establishes that the comparable Cox benchmark is an approximate comparable specification (broad departure from proportionality classification), not a fully assumption-clean Cox model. The comparable Cox still provides the classical methodological anchor for the static early-window family, but all results from this branch should be narrated with the broad-departure approximation explicitly acknowledged. DeepSurv shares the Cox-type ranking structure, but it was not subjected to an identical classical proportional-hazards diagnostic.

\subsubsection*{PH scope boundary}

Table~\ref{tab:appendix_ph_scope_boundary} makes the scope boundary explicit for all nine Family~B models. Formal classical proportional-hazards auditing (via Schoenfeld residuals and log-log diagnostic plots) requires a linear predictor with well-defined martingale residuals; this condition is satisfied only by the comparable Cox anchor, which received a formal audit (broad departure from proportionality classification). For families that use an accelerated failure time or discrete-time formulation (Weibull AFT, XGBoost AFT, Neural-MTLR, DeepHit), the proportional-hazards assumption does not apply and the audit is not applicable. For non-parametric or tree-based ensembles that make no distributional assumption (RSF), the audit is likewise not applicable. For the Gradient-Boosted Cox family, which uses a partial-likelihood objective that implies a proportional-hazards structure, a Schoenfeld audit would be applicable in principle but was not executed within this benchmark freeze. For Royston-Parmar, which relaxes PH via splines, dedicated goodness-of-fit tools exist but were not applied. DeepSurv shares the Cox-type ranking structure but its nonlinear risk surface makes classical Schoenfeld diagnostics inapplicable; this is narrated as a boundary on diagnostic coverage, not as an automatic invalidation of DeepSurv's benchmark performance. Specifically, DeepSurv extends the Cox partial likelihood with a neural predictor $h_0(t)\exp(f_{\theta}(\mathbf{x}))$, where $f_{\theta}$ is a multi-layer network. Schoenfeld residuals require per-subject score contributions derived from the partial-likelihood gradient, which are not available in closed form when the linear predictor is replaced by a neural function; classical proportional-hazards diagnostics are therefore methodologically inapplicable to DeepSurv, not merely omitted. DeepSurv is evaluated through the same predictive metrics (IBS, time-dependent concordance, horizon-specific Brier scores), calibration gaps, and enrollment-level bootstrap rank stability applied to all other families; these constitute the applicable scope of performance evidence for a Cox-type neural model.

\begin{table}[!t]
\caption{Proportional-hazards scope boundary across all nine Family~B models. Formal classical PH diagnostics (Schoenfeld residuals) require a linear partial-likelihood estimator; this condition is met only by the Cox anchor, which was formally audited.}
\centering
\scriptsize
\setlength{\tabcolsep}{4pt}
\renewcommand{\arraystretch}{1.08}
\begin{tabularx}{\linewidth}{@{}>{\raggedright\arraybackslash}p{2.3cm} >{\raggedright\arraybackslash}p{2.0cm} >{\raggedright\arraybackslash}p{1.8cm} X@{}}
\toprule
Model & Assumption class & Formal PH audit & Paper claim boundary \\
\midrule
Cox Comparable & PH (linear) & yes & broad departure from proportionality; all comparable Cox results should be narrated as from an approximate benchmark specification, not from a fully assumption-clean Cox model. \\
DeepSurv & Cox-type neural (nonlinear) & no & Classical Schoenfeld diagnostics require per-subject partial-likelihood score contributions, which are not computable in closed form when the linear predictor is replaced by a neural network. PH diagnostics are therefore methodologically inapplicable (not merely omitted). Evaluated via IBS, concordance, calibration gaps, and bootstrap rank stability. \\
Random Survival Forest & tree-survival (non-parametric) & N/A & No PH assumption is made; PH audit is not applicable. \\
Gradient-Boosted Cox & PH (partial likelihood) & in principle; not run & Partial-likelihood objective implies PH; a Schoenfeld audit would be applicable but was not executed in this benchmark. \\
Weibull AFT & AFT (parametric) & N/A & AFT model; PH assumption does not apply; accelerated time parametrization is used instead. \\
Royston-Parmar & PH spline (relaxed PH) & no & Spline-based relaxation of PH; dedicated diagnostics exist but were not applied. Narrate as relaxed-PH, not audited-PH. \\
XGBoost AFT & AFT (tree-based) & N/A & AFT formulation; PH assumption does not apply. \\
Neural-MTLR & discrete-time & N/A & Discrete-time model; no PH assumption is made or audited. \\
DeepHit & discrete-time & N/A & Discrete-time model; no PH assumption is made or audited. \\
\bottomrule
\end{tabularx}
\setlength{\tabcolsep}{6pt}
\renewcommand{\arraystretch}{1.0}
\label{tab:appendix_ph_scope_boundary}
\end{table}

\subsubsection*{Early-window sensitivity}

Table~\ref{tab:appendix_window_sensitivity} reports IBS and time-dependent concordance for each Family~B tuned family at the canonical window ($w = 4$ weeks), validated against a sensitivity grid ($w \in \{2, 4, 6, 8, 10\}$ weeks). In the pipeline sensitivity grid, performance generally improved with longer observation windows; $w = 4$ was selected as the canonical choice and held fixed for all primary benchmark comparisons.

\begin{table}[!t]
\caption{Early-window sensitivity for Family~B. IBS (lower is better) and time-dependent concordance (higher is better) for each model, reported at the canonical window ($w=4$ weeks). Performance generally improves with longer observation windows; full sensitivity results are available in the pipeline outputs.}
\centering
\small
\setlength{\tabcolsep}{5pt}
\renewcommand{\arraystretch}{1.08}
\begin{tabular}{@{}>{ \raggedright\arraybackslash}p{5.0cm} c c@{}}
\toprule
Model & IBS & TD Concordance \\
\midrule
Random Survival Forest & 0.1118 & 0.6735 \\
Neural-MTLR & 0.1131 & 0.6569 \\
DeepSurv & 0.1145 & 0.6619 \\
Gradient-Boosted Cox & 0.1148 & 0.6621 \\
DeepHit & 0.1151 & 0.5835 \\
Cox Comparable & 0.1176 & 0.6512 \\
Weibull AFT & 0.1180 & 0.6131 \\
Royston-Parmar & 0.1190 & 0.6430 \\
XGBoost AFT & 0.1495 & 0.5772 \\
\bottomrule
\end{tabular}
\setlength{\tabcolsep}{6pt}
\renewcommand{\arraystretch}{1.0}
\label{tab:appendix_window_sensitivity}
\end{table}

\FloatBarrier

\subsection*{Reliability diagrams}

\begin{figure}[!ht]
\centering
\includegraphics[width=.80\linewidth,height=.58\textheight,keepaspectratio]{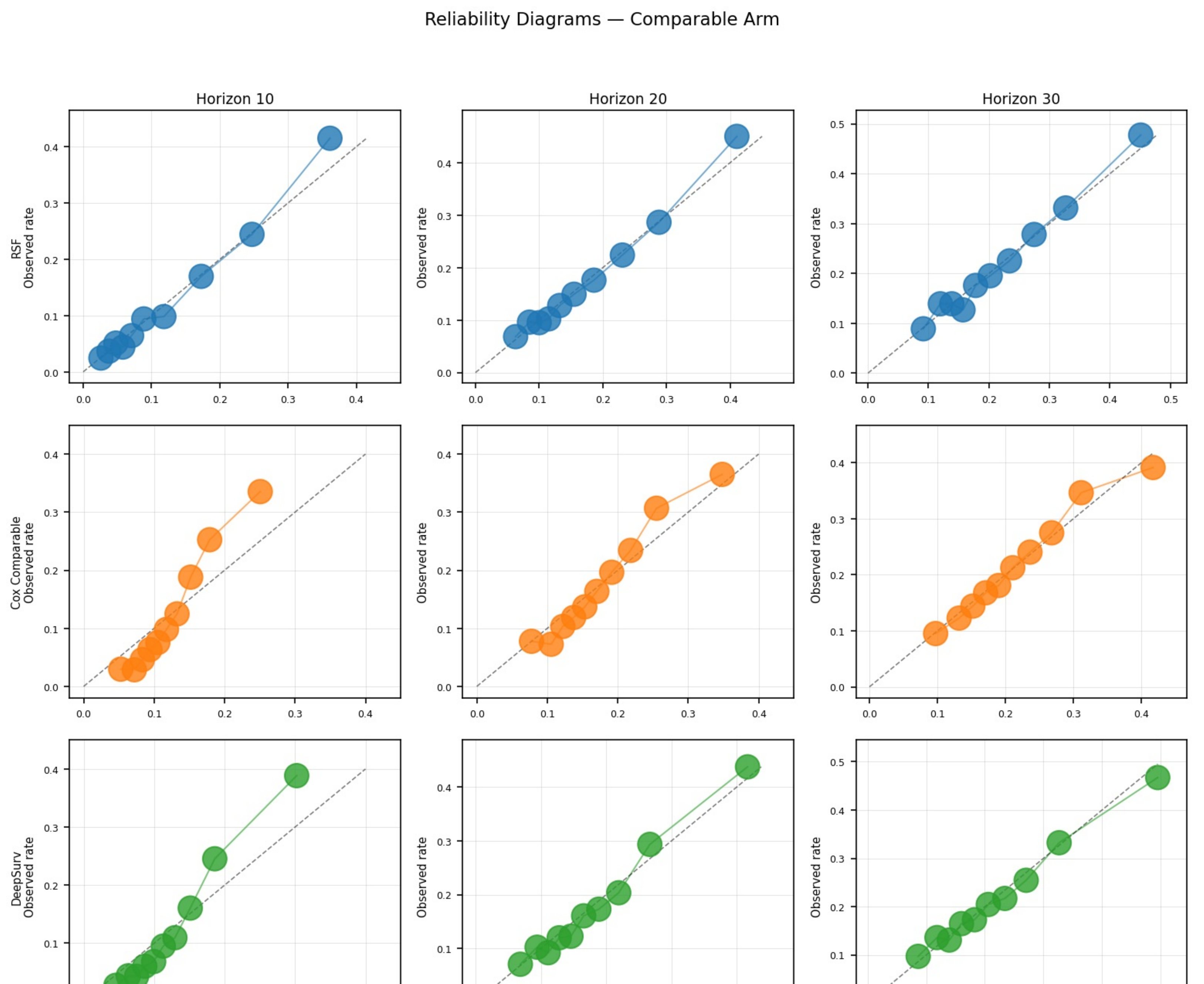}
\caption{Reliability diagrams for Family~B (panel 1 of 3). This panel covers the first subset of static early-window families; most points remain close to the diagonal, consistent with the strong calibration profile of the leading static early-window models.}
\label{fig:appendix_reliability_comparable_a}
\end{figure}

\begin{figure}[p]
\centering
\includegraphics[width=.80\linewidth,height=.58\textheight,keepaspectratio]{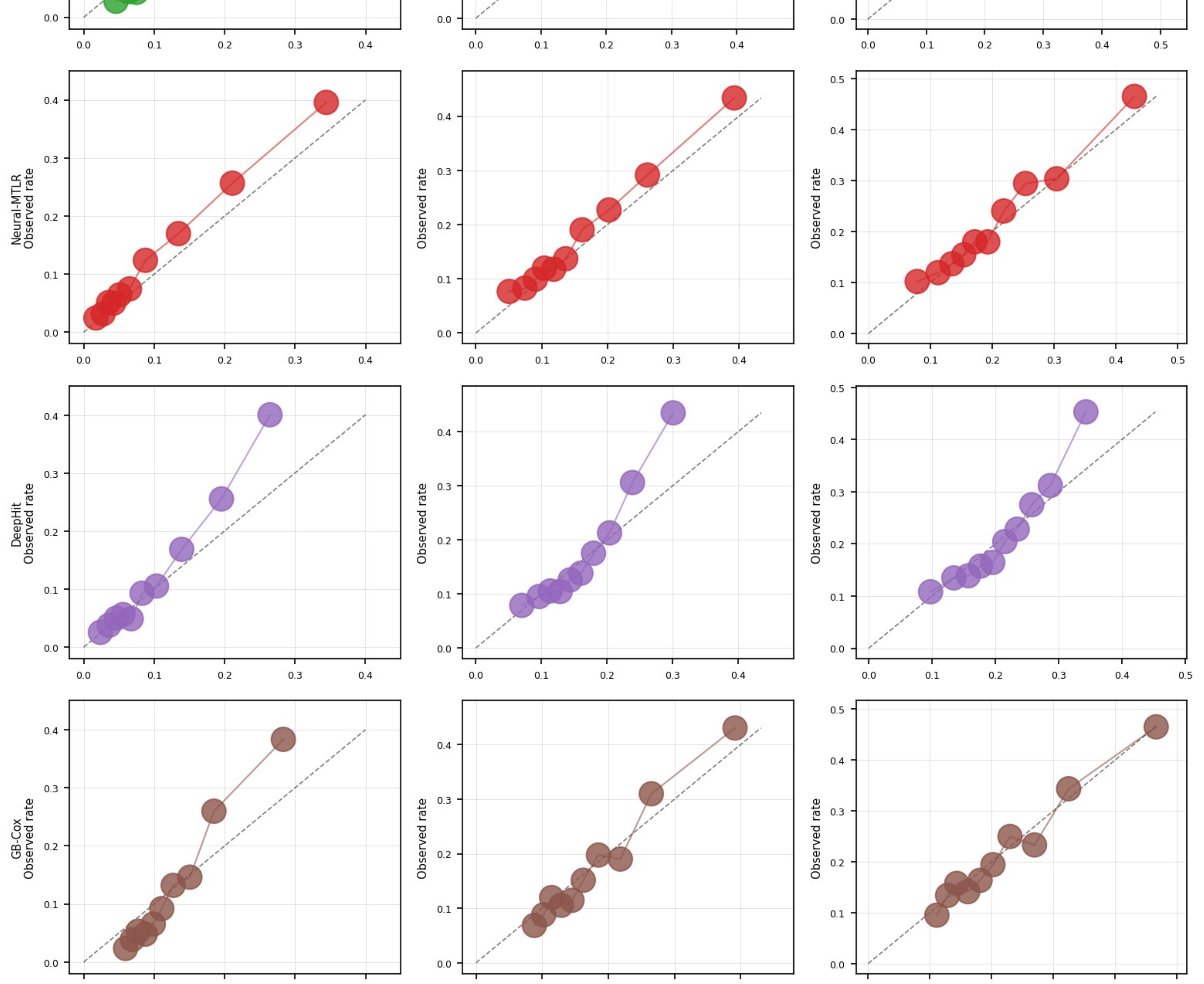}
\caption{Reliability diagrams for Family~B (panel 2 of 3). The middle subset preserves the same three-horizon calibration comparison while avoiding the page overflow of the original full-width grid.}
\label{fig:appendix_reliability_comparable_b}
\end{figure}

\begin{figure}[p]
\centering
\includegraphics[width=.80\linewidth,height=.58\textheight,keepaspectratio]{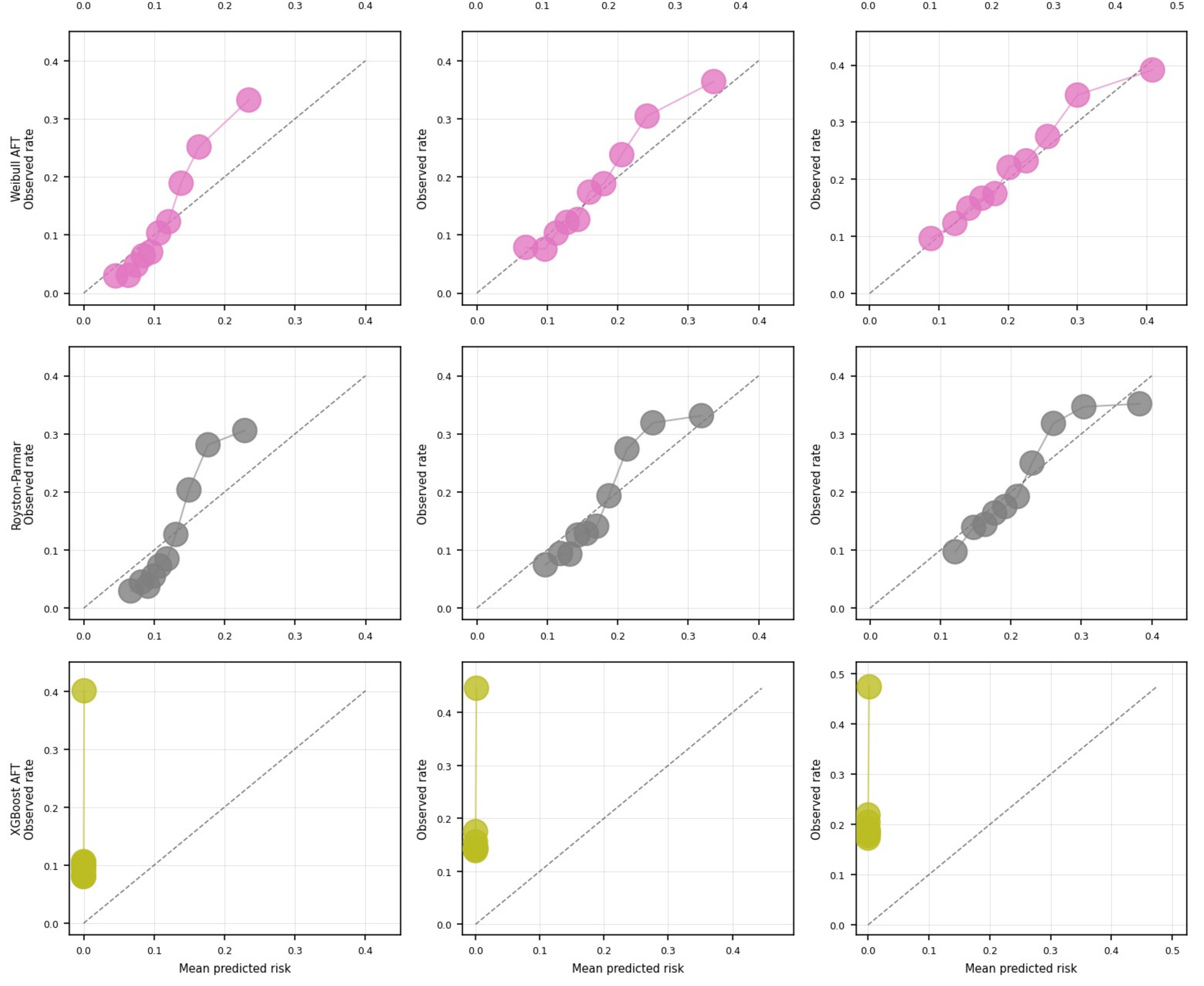}
\caption{Reliability diagrams for Family~B (panel 3 of 3). This final panel includes the remaining static early-window families and retains the visible calibration departure of XGBoost AFT relative to the better-calibrated alternatives.}
\label{fig:appendix_reliability_comparable_c}
\end{figure}

\begin{figure}[!ht]
\centering
\includegraphics[width=.9\linewidth]{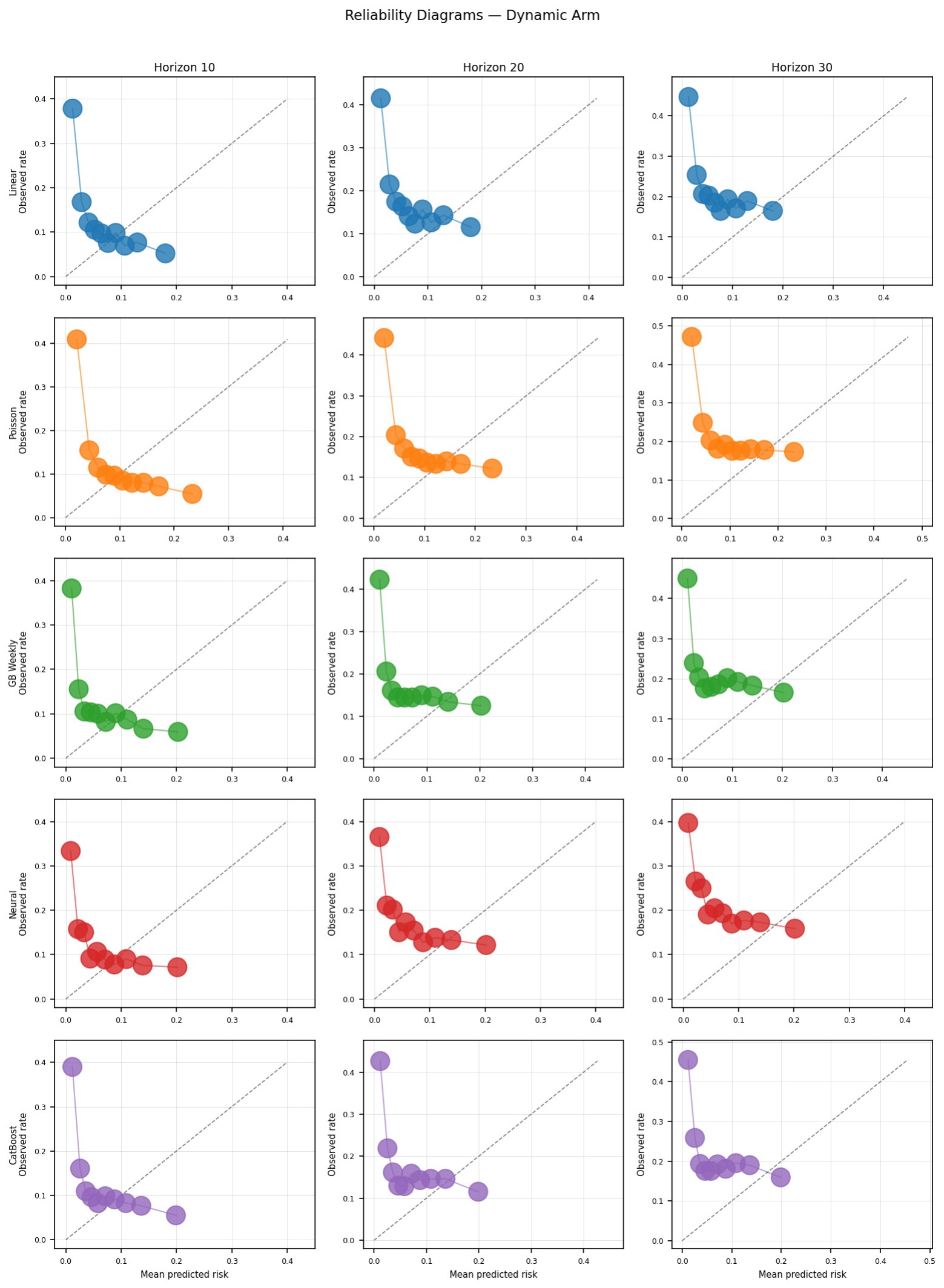}
\caption{Reliability diagrams for Family~A. Rows correspond to horizons 10, 20, and 30~weeks; columns correspond to models. Family~A calibration gaps remain wider than those of the strongest Family~B models, but the panels still show coherent horizon-wise probability structure.}
\label{fig:appendix_reliability_dynamic}
\end{figure}

Figures~\ref{fig:appendix_reliability_comparable_a}, \ref{fig:appendix_reliability_comparable_b}, and~\ref{fig:appendix_reliability_comparable_c} display the reliability diagrams for Family~B in three smaller panels to preserve page layout. Figure~\ref{fig:appendix_reliability_dynamic} retains the Family~A reliability view. In every panel, rows correspond to horizons 10, 20, and 30~weeks; each subplot compares mean predicted risk (x-axis) against observed event rate (y-axis), with point size proportional to bin size and the dashed diagonal representing perfect calibration. These figures complement the tabulated calibration evidence in Table~\ref{tab:calibration_summary} and Appendix~\ref{sec:calib_evidence}.

\FloatBarrier

\end{appendices}

\backmatter

\bmhead{Funding}
This research received no external funding.

\bmhead{Data availability}
The datasets analysed during the current study are available in the
Open University Learning Analytics Dataset (OULAD) repository,
\url{https://research.stem.open.ac.uk/ouanalyse/dataset/}.
Direct download at \url{http://schools.stem.open.ac.uk/cdn/files/anonymisedData.zip}
\citep{Kuzilek2017OULAD}.

\bmhead{Ethics, Consent to Participate, and Consent to Publish declarations}
Not applicable. This study uses the publicly available Open University Learning Analytics Dataset (OULAD), which contains only anonymized records. No human subjects were recruited, no identifiable personal data were collected, and no intervention was administered. Accordingly, ethics approval, consent to participate, and consent to publish were not required.

\bmhead{Supplementary information}

Online Resource~1 (\texttt{ESM\_1.pdf}): hyperparameter search grids for
all 14~models, including candidate count, selection criterion, and full
grid specification.

\bibliography{bib}

\end{document}